# On the Control of Attentional Processes in Vision

*John K. Tsotsos, Omar Abid, Iuliia Kotseruba, Markus D. Solbach*
*York University, Toronto, Canada*
*Contact: tsotsos@eecs.yorku.ca*

## Abstract

The study of attentional processing in vision has a long and deep history. Recently, several papers have presented insightful perspectives into how the coordination of multiple attentional functions in the brain might occur. These begin with experimental observations and the authors propose structures, processes, and computations that might explain those observations. Here, we consider a perspective that past works have not, as a complementary approach to the experimentally-grounded ones. We approach the same problem as past authors but from the other end of the computational spectrum, from the problem nature, as Marr's Computational Level would prescribe. What problem must the brain solve when orchestrating attentional processes in order to successfully complete one of the myriad possible visuospatial tasks at which we as humans excel? The hope, of course, is for the approaches to eventually meet and thus form a complete theory, but this is likely not soon. We make the first steps towards this by addressing the necessity of attentional control, examining the breadth and computational difficulty of the visuospatial and attentional tasks seen in human behavior, and suggesting a sketch of how attentional control might arise in the brain. The key conclusions of this paper are that an executive controller is necessary for human attentional function in vision, and that there is a 'first principles' computational approach to its understanding that is complementary to the previous approaches that focus on modelling or learning from experimental observations directly.

## 1.0 Introduction

There are a few basic realities that play a major role towards the key conclusions of this paper. Any brain or behavioral process takes time. Each computation takes time, each transfer of information from one representation to the next takes time because information must travel over some neural distance and at a speed determined by the type of connection, each motor action takes time, and so on. Each behavior has a specific time course and its component visual sub-tasks each require different actions, different timings of actions, different durations, etc. It feels immediate to us when we look at a photo of a single face and detect that it is our mother, yet it will necessarily take much longer to find that same face in a crowd. The former task may not require eye movements yet the latter may need many, each requiring about 250ms to set up and execute. The number of different kinds of visual behaviors humans can execute seems unbounded and the variability of both time to achieve and success of execution equally variable.

As we will summarize later, many have considered the problem of executive or attentional control. We approach the same problem as past authors but from the other end of the computational spectrum, from the general and abstract to the detailed and specific, as Marr's three levels would prescribe (Computational, Representational and Algorithmic, Implementation - Marr 1982). Here we address only the most general level, the computational level: What problem must the brain solve when orchestrating attentional processes in order to successfully complete one of the many possible visuospatial tasks at which we as humans excel? In Tsotsos (1984) Marr's 3 levels were re-framed as Content, Form and Use, while in Tsotsos (1987) a fourth level, the Complexity Level was added. This will play a foundational role throughout this work: the requirement that any model or theory must be shown to be realizable under the constraints of the resources available and under the performance specifications that any realization must exhibit (Tsotsos 2017).

Consider the simple act of seeing a face in an image. Suppose you are a subject in a perceptual experiment and the experimenter is sitting across a table from you. You are told that the experimenter will hold up a picture for you to see and in advance of it, you will be asked a question about the picture which you should answer as quickly as possible. The picture will be shown as long as you have not replied, but once you reply, it will be put down. What is the sequence of actions within your visual system that are in play for this task? First of all, the initial instructions you receive set you up to expect a picture and to receive a question. The question sets up your visual system to expect something; say it's "do you see your mother's face?". Whatever might be irrelevant to this task might be suppressed or ignored by your visual system. From an efficiency point of view, the irrelevant should be ignored to conserve resources. Then the picture is held up, you detect its location and move your gaze to fixate the picture, perhaps in the center as a default. The resulting image is processed by your expectant visual system. It is a picture of single face on a background depicting a dining room. Presumably, some decision process would realize that the neurons that represent your mother's face are strongly firing and would then signal you to respond positively to the question posed. The picture is taken away and your gaze moves to something else in anticipation of the next picture, perhaps the experimenter's face. Your visual system can't help but process what it sees, so new face features now are the more active ones in your visual system. Another question is asked in advance of the next image; say it's to detect your father's face, and you re-tune your visual system to prepare for this. This re-tuning might instruct the system to no longer expect your mother's face, either in terms of its size, features or location and impose the characteristics of your father's face as the relevant ones, suppressing others. You also prepare to disengage from viewing the experimenter's face. The next picture is shown, you detect its location, disengage from the experimenter's face, plan for a saccade, and you orient your gaze to it. This time, it is not a single face on a known background but a group of people in an outdoor scene where one of them might be your father. The same process, which has now been primed as the one to use for detection and decision-making is no longer the right one. The face is now much smaller, and is present in a conflicting background because there are many faces and they all share some similarities simply by being faces. You decide to begin a search but this requires you first to re-tune your visual system to process the smaller faces. You locate a face and then move your eyes to it and scrutinize the small features to check if that face could be your father's. You move from one to the next until you look at them all. If you cannot find your father, it is likely you might go back and check some of the faces just in case you made a mistake, perhaps the lighting or viewpoint led to mis-leading cues. Eventually you will decide whether or not you detect your father's face. This is a much longer process and one that requires eye movements and all the processing they entail.

One more example should cement the point. Suppose you take your 12-yr old daughter, who is an aspiring dancer, to a dance workshop. You return to retrieve her, open the door of the class and are confronted with a hundred young girls in a large hall, milling about and chatting or playing as they wait for their parents, all wearing exactly the same clothes (their dance outfit), all with their hair put up in a bun, all roughly the same height and age, and surprisingly most girls have the same skin and hair colour as your daughter. You take a quick look at the crowd and begin to panic that you can't see your daughter. The familiarity you have with your daughter's face does not help; she might not even be facing you. You switch to a search strategy and quickly scan the faces. You still have no luck. You then move to a different strategy and move about looking at faces from closer distances as well as moving to see faces not oriented towards your line of sight. Of course, you eventually find your daughter and all is well. But the real question for us is, how did you do this? What sequence of actions, perceptions, decisions must have taken place in your brain for you to complete the finding of your daughter? A major difference between this example and the previous ones is the need to be an active observer, the need to select what image to process at each time instant and to move to acquire it in support of your overall goals. These examples suffice to argue for why some kind of controlling process is needed. The important questions are how are all these actions and decisions taken at the right time, in the right order, and monitored for their correctness?

The attention literature has been dominated by resource concerns: many have written about how attention allocates resources but what exactly these resources might be and how an allocation is achieved has received far less consideration. As appears in Tsotsos (2011), "notions of resource allocation persist in the literature even though there is no evidence whatsoever for any physical counterpart; no neurons can be physically moved around, no connections can be physically moved dynamically (although strong modulations of neurons or connections may show some, but not all, of the characteristics or such a physical reallocation), and no visual areas can, in a dynamic real-time fashion, change to become auditory and then change back again." The hardware of the brain seems fixed - the same brain is used for all behaviors - and is not rapidly altered save for strengthening or weakening components temporarily. This leads to the definition of attention presented in Tsotsos (2011): *Attention is the process by which the brain controls and tunes information processing*. This tuning is one form of resource allocation: it specializes the brain's machinery from generic to specific for the task and stimuli of the moment and then sets it back again. Specializing the brain machinery has important impacts because it causes the brain to be less susceptible to noise and conflicting information, more efficient in its processing since search spaces are reduced, and, sharpened in its decision-making stages because interfering stimuli and knowledge are limited to those most relevant to the current task. The other resource is time. The above examples should make clear that different behaviors take different amounts of time to complete. It could be that exactly the same machinery is used in exactly the same manner for all behaviors, and that for difficult cases, that machinery simply takes longer to "settle" on an answer, perhaps using some kind of slow information accumulation. This is unlikely as will be discussed later. It seems far more likely that the amount of time used for a given task is dependent on the time when the task must complete, the sub-tasks that must be completed and their order, and this implies an active time management process.

It would not be unreasonable to suppose that you know how to search a picture for a target because over time you have learned the algorithm. Such an algorithm for a complex visual behavior has been termed a Cognitive Program (Tsotsos & Kruijne 2014) or Visual Routine (Ullman 1984). Ullman's proposal addressed how human vision extracts shape and spatial relations. He proposed that: visual routines (VRs) compute spatial properties and relations from base representations to produce incremental representations; VRs are assembled from elemental operations; new routines can be assembled to meet processing goals; different VRs share elemental operations; a VR can be applied to different spatial locations; mechanisms are required for sequencing elemental operations and selecting the locations at which VRs are applied; and, VR's can be applied to both base and incremental representations. VRs have been studied and supported by both computer scientists (Johnson 1993; McCallum 1996; Horswill 1995; Brunnström et al. 1996; Rao 1998) and brain scientists (Roelfsema et al. 2000, 2003; Roelfsema 2005; Cavanagh et al. 2001) and seem an enduring idea. However, the idea is dated in its detail and requires updating. Cognitive Programs (Tsotsos 2010; Tsotsos & Kruijne 2014) are an elaboration and modernization of the original idea and are based on a broader up-to-date view of visual attention and visual information processing. Cognitive Programs (CPs) are sequences of representational transformations, attentional tunings, decisions, actions, and communications required to take an input stimulus and transform it into a representation of objects, events or features that are in the right form to enable solution of a task. In a very real sense, CPs are algorithms for solving problems (see Lázaro-Gredilla, et al., 2019, for a practical example).

It might be that you have learned thousands of such cognitive programs (CP) and you have them stored in memory, quickly deploying the right one at the right time, dynamically parameterized for the task. These CPs would provide an encoding of algorithms for visual behaviors, including attentional actions. Many attentional mechanisms are easily evident in the examples given earlier: covert attention, disengage attention, engage attention[1], inhibition of return, neural modulation, overt attention, priming, recognition, search, selection, shift attention, and visual working memory. Generally, the timings of these actions need synchronization and coordination. The CP algorithm for a particular visual behavior represents an ordered and parameterized sequence of actions comprised of these attentive elements, in addition to the other necessary computations.

One might imagine that even a large set of CPs might be in memory and can proceed without any oversight by a controller. This might be valid if all CPs are always successful as specified. This is clearly untrue and some process is needed to monitor the execution of a CP, detect when it fails, adjust the program or try a new one (recall the dance workshop example) until success is achieved.

Since 1987, we have been developing the Selective Tuning model of visual attention (ST), demonstrating its computational elements and gathering supporting human experimental evidence for its key predictions (for a history of its development, see Tsotsos 2011). By all counts, ST can be considered a strong contender for how human visual attention might function (e.g., Carrasco 2011, Herzog & Clarke 2014). Naturally, we continue to investigate its falsifiability, and to do so it must be extended so that its behaviour and predictions can be tested in the larger context of an agent that exhibits broader perceptual and action performance. It seems straightforward that the embedding of ST within a larger behavioral and cognitive setting would involve how it interacts with the control of eye fixations, how it interacts with the behavioral or cognitive decision-making for the task being performed, and how it interacts with the various memory systems required for the task. This larger context has led us to the STAR cognitive architecture (**S**elective **T**uning **A**ttentive **R**eference; first described in Tsotsos & Kruijne (2014)). The principal components of STAR are the Task Executive, the Attention Executive, Selective Tuning, Working Memory, Perception Hierarchy, and Fixation Control. STAR was not designed with the same goals (general purpose intelligence) as other well-known cognitive architectures such as SOAR (Laird et al. 2017, Kotseruba & Tsotsos 2020). Rather, our goal is to consider a subset of the goals of such architectures, focussing on the details of how attention functions, broadly considered, for vision and behaviour (and not on abilities such as natural language processing, general-purpose planning and reasoning, etc.). The goals of this paper are driven strongly by the need to understand how attention connects to the other elements of STAR.

## 2.0 A Selective Evolution of Views on Executive Control

As one looks over the literature related to this topic, it quickly becomes bewildering, with contributions spanning 5 decades at least from multiple disciplines. How can all of this be reconciled into some kind of single view of the best current understanding? Here, we will admit that it likely cannot. This is not because of any flaws or errors; rather, there is little consistency with respect to the level of abstraction[2] being addressed and most works address more than one while not being clear about which ones.

A detailed review of the many approaches and opinions regarding attentional control would be beyond the scope of this paper. What follows is a brief and selective view of the relevant previous concepts, with the contributions chosen to be representative of the past 50 years as well as the many disciplines involved. There is a distinct selection bias for works that address the issues central to this paper. Figure 1 suggests a taxonomy of research efforts. There are five major classes and each is now defined. Some papers are descriptive; they contribute important principles that may underlie attentional control, for example. This class is denoted by DS and will be used for papers that are purely descriptive. It is assumed that all papers in the other classes also contribute to basic ideas and principles and thus all the other classes are specializations of the DS class. Directed graph models (DG) are those described primarily any graphical node-and-arrow depiction of elements of control (anatomical, neural, functional). They include efforts to determine which brain areas are involved. Functional models (FN) attempt to provide a specification of the input-to-output mapping of stimulus to behavior. Mathematical models (MA) propose mathematical descriptions of control, specifying variables of relevance and their precise relationships. Finally, algorithmic models (AL) give a step-by-step proposal of how control is implemented. The assignments to the classes are naturally subjective and criteria are applied generously.

---

[1] Within Selective Tuning, attention may be focussed on a spatial location, to a feature, to an object or event, or a combination as is appropriate for the current task. As a result, the terms *engage* and *disengage* apply to to the current attentional focus.
[2] The many levels of abstraction relevant for theories of the brain have been discussed elsewhere (e.g., see Churchland & Sejnowski 1993 for a neuroanatomical set of levels, Ballard 2015 for a computational set of levels).

We suggest that a complete model should provide details at each of these levels of description. In comparison with the levels of abstraction seen in Churchland & Sejnowski (1993) or Ballard (2015), this seems a compressed decomposition and further makes no direct mention of the physical reality of neural structures (e.g., Churchland & Sejnowski give the levels as molecules, synapses, neurons, networks, maps, systems, CNS). This would require an orthogonal taxonomy because each of the 5 classes here could be considered at each of those levels, and is beyond our scope.

Further, we note that previous efforts mostly do not deal with the mechanistic (Brown 2014) issues relevant to the problem. Borrowing from our own implementations of our attentional model, it is evident that synchronization, initiation, termination, monitoring, comparison, communication and decision-making are the minimum set of such mechanistic issues. Further, most past works do not not differentiate between data and control connections, nor do they deal with the realities of the formal computational problem as will be described below. That is, to them, the problem of "vision" or "visuospatial behavior" is a monolith. Later, it will be shown that this is not the case and that the decomposition of the problem matters greatly in how a solution strategy may be developed. The brief and selective sampling of this past research follows.

Kahneman (1970) in a pioneering paper makes several suggestions based on experiment and introspection for attentional control characteristics. For instance, he points out that spatial orientation appears to be a common mode of attentional control, that selective attention is far from optimal in selective monitoring and in selective listening without response, in contrast to shadowing, and that an important determinant of effective attentional control is the need for continuous, coherent, serially organized behavior. Allport (1993), on the other hand, believed that it is pointless to focus on the "locus of selection" and on "which processes do and do not require attention". "Attentional functions are of very many different kinds, serving a great range of computational functions. There can be no simple theory of attention." The conclusions of the current paper fit well with this view.

For Posner & Dehaene (1994), the term 'executive' suggests two important overall functions. First, an executive is informed about the processes taking place within the organization, and second, an executive exercises some control over the system. They argue for the role of the anterior cingulate in both regards. By contrast, Singer & Gray (1995) propose a temporal correlation hypothesis. This hypothesis asserts that synchronization of neuronal activity on a millisecond time scale may be exploited to link featural information that is represented in different parts of the cortex. Synchronization can, in principle, be used to select with high spatial and temporal resolution those activity patterns that belong together and to enhance the effect of this activity so it may be evaluated for further processing.

Egeth and Yantis (1997) in a classic paper identify the key issues as being: control of attention by top-down (or goal-directed) and bottom-up (or stimulus-driven) processes; representational basis for visual selection, including how much attention can be said to be location or object based; and, the time course of attention as it is directed to one stimulus after another. They are certainly correct on all points and it is unfortunate that of these, the time course issues seem least developed in the intervening decades. Dehaene, Kerszberg & Changeux (1998) propose the Global Computational Workspace model that permits mutual connectivity and thus reconfiguration among the many processes of the brain, specifically, long-term memory, attention, perception, motor, and value systems. Using it, they build a simulation network that shows how the Stroop Task might be accomplished. Yantis (1998) reminds us of the important distinction made by James (1890) that attention is *active* when goal-driven (driven by the observer's deliberate intentions and strategies) whereas if driven by some salient aspect of the stimulus, not necessarily associated with the observer's goals, attention is *passive*. These terms are preferred to their more common counterparts, top-down and bottom-up, because they are more specific and explicitly include the observer. He then provides a wealth of evidence that demonstrates how any attentional act is some combination of the two and these might work in concert. Here, we certainly agree with this active-passive distinction (see Bajcsy et al. 2019 for the computational counterpart). Active perception involves not only eye movements, but also head and body movements, Any behavior by a subject that facilitates the perception of the stimuli needed for the current task is an instance of active perception (see a nice overview in Findlay & Gilchrist 2003).

Normal visual behavior is accomplished through a continuous cycle of fixation and visual analysis interrupted by saccades according to Schall & Bichot (1998). During fixation of one point in the image at least three processes take place: the visual analysis of the image centered on the fovea to ascertain its identity; the visual analysis of the image in the periphery to locate potential targets for the next saccade; and, the production of the saccade. They highlight important questions, namely, how much time does perceptual processing and response preparation take? Can these two processes overlap? How late can perceptual processing influence response preparation? Overall, they take a serious look at the elements that comprise the overall time course of saccades. Carpenter (2000) provides a schematic representation of some of the main features of the saccadic control system and the brain areas that are involved. Many other authors also provide perspectives on which brain areas have involvement for a variety of tasks (Corbetta & Shulman 2002; Fuster 2002; Cole & Schneider 2007; Baluch & Itti 2011; Hopfinger, Buonocore & Mangun 2000). These are no doubt useful; however they are not sufficient to delineate a solution to the control problem. Gilbert & Li (2013) describe the large set of re-entrant or feedback pathways between cortical areas that carry information about behavioral context, including attention, expectation, perceptual task, working memory and motor commands. They suggest that neurons receiving such inputs effectively function as adaptive processors that are able to assume different functional states according to the task being executed. They emphasize the dynamic nature of the receptive field, which allows neurons to carry information that is relevant to the current perceptual demands. They provide additional support for our definition of attention (Tsotsos 1990; Tsotsos 2011).

Giesbrecht, Woldorff, Song & Mangun (2003), Roelfsema (2005) and Niendam, Laird, Ray, Dean, Glahn & Carter (2012) contribute to a problem decomposition idea which will be seen as central to our work as well. Giesbrecht, Woldorff, Song & Mangun (2003) show that top-down control systems operate by generalizing across spatial and nonspatial dimensions as well as by recruiting specialized systems. In other words, there is a decomposition of function that is productively used by top-down control. Roelfsema, Khayat & Spekreijse (2003) and Roelfsema (2005) lay the neural foundation of visual routines or cognitive programs by showing how visual search, cuing, region filling, working memory, matching, and more can assembled into routines and coordinated in time to perform different tasks, as revealed by single cell recordings. In a similar vein, Niendam, Laird, Ray, Dean, Glahn & Carter (2012), using fMRI, show that the attentional functions of switching tasks, inhibitions, working memory, initiation, planning and vigilance are continually recruited within a frontal–cingulate–parietal–subcortical cognitive control network in the performance of a wide variety of cognitive tasks. VanRullen (2016) points out that there are several rhythms of perception that may depend on sensory modality, task, stimulus properties, or brain region. In vision, for example, a sensory alpha rhythm (~10 Hz) may coexist with at least one more rhythm performing attentional sampling at around 7 Hz. In other words, attention has a cyclic nature something that was hinted at in Tsotsos et al. (1995). Mansouri et al. (2016) suggest that conflict might be one among many drivers of adjustments in executive control and that the ACC might be just one component of overlapping distributed systems involved in context-dependent learning and behavioral control.

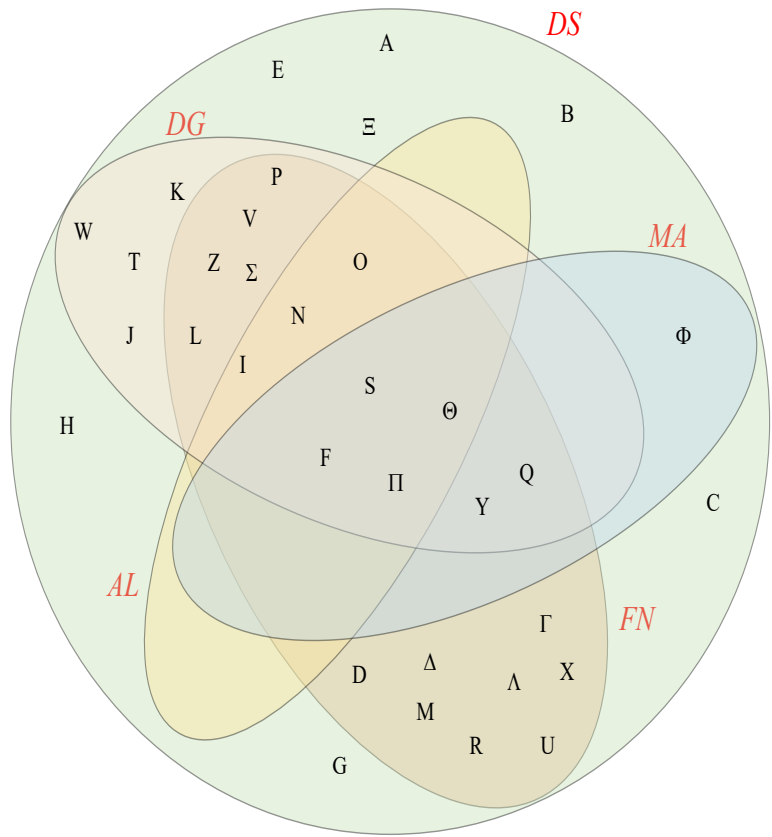

**Figure 1.** A Taxonomy of Attentional Control Literature. Papers that address attentional control are grouped into 5 major classes (class labels in red): descriptive (DS); directed graph models (DG); functional models (FN); mathematical models (MA); and, algorithmic models (AL) (defined above). Paper IDs are placed on the Venn diagram depending on which kinds of contributions are made, to the best of our understanding and judged as generously as possible. The list of papers included is certainly incomplete, but intended to provide a reasonable sampling of research, of different types, over the past 5 decades.

| ID | Reference |
|---|---|
| A | Kahneman 1970 |
| B | Allport 1993 |
| C | Posner & Dehaene 1994 |
| D | Singer & Gray 1995 |
| E | Egeth & Yantis 1997 |
| F | Dehaene, Kersberg & Changeux 1998 |
| G | Schall & Bichot 1998 |
| H | Yantis 1998 |
| I | Carpenter 2000 |
| J | Hopfinger, Buonocore & Mangun 2000 |
| K | Corbetta & Shulman 2002 |
| L | Fuster 2002 |
| M | Giesbrecht, Woldorff, Song & Mangun 2003 |
| N | Roelfsema, Khayat & Spekreijse 2003 |
| O | Roelfsema 2005 |
| P | Cole, Schneider 2007 |
| Q | Gold & Shadlen 2007 |
| R | Rossi et al. 2009 |
| S | Zylberberg, Slezak, Roelfsema, Dehaene & Sigman 2010 |
| T | Baluch & Itti 2011 |
| U | Niendam, Laird, Ray, Dean, Glahn & Carter 2012 |
| V | Cole, Reynolds, Power, Repovs, Anticevic & Braver 2013 |
| W | Gilbert & Li 2013 |
| X | Miller & Buschman 2013 |
| Y | Womelsdorf & Everling 2015 |
| Z | Desrochers, Burk, Badre, & Sheinberg 2016 |
| Γ | Mansouri, Egner, Buckley 2016 |
| Δ | Van Rullen 2016 |
| Λ | Hanks & Summerfield 2017 |
| Θ | Laird, Lebiere & Rosenbloom 2017 |
| Π | Abid 2018 |
| Σ | Batista-Brito, Zagha, Ratliff & Vinck 2018 |
| Φ | Tang & Bassett 2018 |
| Ξ | Moyal & Edelman 2019 |

At a finer level of abstraction than brain area, Womelsdorf & Everling (2015) present a set of neural configurations (or motifs) for specific attentional functions or circuits including disynaptic disinhibition, dendritic gating, and feedforward inhibitory gain control. They claim that these common circuit motifs are used to coordinate attentionally selected information across multi-node brain networks during goal-directed behavior. Abid (2018) demonstrates an implementation of this neural motif idea of Womelsdorf and Everling, which he terms Neural Primitives within the STAR architecture, using Cognitive Programs successfully showing performance on a variety of tasks, including Localization, Priming, Detection, Recognition, Categorization Identification.

Moyal & Edelman (2019) discuss the benefits of optimization for recurrent networks while Tang & Bassett (2018) provide a tutorial-like overview of how control theory may be productively used by neural models for implementation of optimization objectives. Optimality criteria and how they may be achieved are important. Gold and Shadlen (2007) describe how neurons in sensorimotor areas, including the parietal and dorsal prefrontal cortex, contribute to perceptual decisions by optimizing input signals through repeated sequential sampling and linear integration to a fixed decision threshold. Mansouri, Egner, Buckley (2016) suggest that conflict might be one among many drivers of adjustments in executive control.

Batista-Brito, Zagha, Ratliff & Vinck (2018) study how contextual modulations of sensory processing are implemented within the local cortical circuit. They propose a hierarchical predictive coding model modified from Rao and Ballard (1998). In their model, prediction signals (feedback) are subtracted from sensory signals (input), and the difference is propagated (feedforward) as a prediction error. This process may be implemented along the feedforward and feedback cortical processing streams. In their view, sensory prediction and top-down attention are two principal types of contextual processing occurring within the cortex: Predictive coding operations encompass the differential encoding of sensory inputs depending on expectancies and priors (i.e. how likely an input is), and the emergence of neurons that encode unexpected sensory prediction errors. Attention entails the modulation of neural activity depending on internal goals (i.e. how relevant an input is). Expectancy and attention may modulate neural activity in a wide variety of ways, from being complementary to opposing

Several authors propose possible control signals. Rossi et al. (2009) show that the PFC plays a critical role in the ability to switch attentional control on the basis of changing task demands. Zylberberg Slezak, Roelfsema, Dehaene & Sigman (2010) developed a spiking-neuron implementation of a cognitive architecture in which the precise sensory-motor mapping relies on a network capable of flexibly interconnecting processors and rapidly changing its configuration from one task to another. Simulations suggest that seriality in dual (or multiple) task performance results as a consequence of inhibition within the control networks needed for precise routing of information flow across a vast number of possible task configurations. Miller and Buschman (2013) specify particular cortical circuits for the control of attention via a frontoparietal network acting on visual cortex, regulated via rhythmic oscillations. Hanks and Summerfield (2017) assert that decisions are not merely driven by accumulation of noisy sensory evidence, but by time-varying bias signals that help curtail deliberation in the face of ambiguous information.

Cole & Schneider (2007) and Cole et al. (2013) describe the frontal-parietal network (FPN) and distinguish it from other brain networks involved in cognitive control (the cingulo-opercular control network, the salience network, the ventral attention network, and the dorsal attention network). They point out that the FPN includes portions of lateral prefrontal cortex (LPFC), posterior parietal cortex (PPC), anterior insula cortex, and medial prefrontal cortex. They propose their Flexible Hub Theory and predict that most adaptive task-control flexible hubs likely exist within the FPN, based on evidence that the FPN is especially active during scenarios requiring highly adaptive task control. They suggest that "some brain regions flexibly shift their functional connectivity patterns with multiple brain networks across a wide variety of tasks. Compositional coding refers to the possibility of a systematic relationship between connectivity patterns and task components (for example, rules), allowing well-established representations to be recombined, and therefore reused, in novel task states to enable transfer of knowledge and skill across tasks. Together, these two mechanisms describe a distributed coding system that provides for an efficient means of implementing a wide variety of task states while also providing the means for novel task control, the ability of humans to quickly perform new tasks based on instruction alone". Desrochers et al. (2016) writes a related perspective: "Our ability to plan and execute a series of tasks leading to a desired goal requires remarkable coordination between sensory, motor, and decision-related systems. Prefrontal cortex (PFC) is thought to play a central role in this coordination, especially when actions must be assembled extemporaneously and cannot be programmed as a rote series of movements. A central component of this flexible behavior is the moment-by-moment allocation of working memory and attention. The ubiquity of sequence planning in our everyday lives belies the neural complexity that supports this capacity, and little is known about how frontal cortical regions orchestrate the monitoring and control of sequential behaviors." These proposals remind one very strongly of the Cognitive Programs described earlier.

Finally, there is a rather large literature on the development of cognitive architectures (reviewed in Kotseruba & Tsotsos 2020) and each necessarily addresses control to some extent because they are all intended to be flexible in purpose. In Laird, Lebiere & Rosenbloom (2017), a proposal is made for a 'standard model'. In this standard model, local control is provided by procedural long-term memory and composed of rule-like conditions and actions, and imposes control by altering contents of working memory. Complex behavior arises from a sequence of independent cognitive cycles that operate in their local context, without a separate architectural module for global optimization (or planning).

In conclusion, there is a wealth of promising directions and ideas represented by these papers and as will become clear many find their way into out thinking here. Particularly, the importance of flexible composition of elements to achieve solutions for dynamic task presentation, is highlighted by several. Even though some present computational simulations with impressive performance, these do not really address a 'first principles' computational approach, and this is the concern of the following sections.

## 3.0 The Nature of Attention Control in Vision

On the one hand 'attention' feels easy, as James (1890) wrote "Everyone knows what attention is. It is the taking possession by the mind, in clear and vivid form, of one out of what seem several simultaneously possible objects or trains of thought." On the other hand, it seems far too complex to approach in a rigorous, formal manner. Excellent past reviews address attention and its control (see Itti, Rees & Tsotsos 2005, Carrasco 2011, Nobre & Kastner 2014), and although agreement at a high level of abstraction is building, consensus at more detailed levels seems evasive and many opinions still pervade the literature. Here, we attempt to ameliorate this problem and begin by asking key questions about attentional control.

First, what exactly is there to control? Any controller needs to impact something, a process, a system of some type, with some purpose to the impact. For our context, a controller would impact how humans react to and solve problems posed by their visuospatial environment. To initiate our computational exploration, we first need to understand the computational nature of these visuospatial problems. As Marr's (1982) computational level of a theory asked: What is the goal of the computation, why is it appropriate, and what is the logic of the strategy by which it can be carried out? Answers to these questions will tell us what there is to control. Second, is the solution of such a visuospatial problem in need of explicit control? At an intuitive level, if a system might need control it means that in some way it is not desirable for it to be 'out of control'. It is not unreasonable to assume that as a rule, human agents are expected to be rational agents. They have models of their world and they behave in a manner that is purposeful and consistent with their models to achieve their goals. In other words, behavior inconsistent within this paradigm means the agent is 'out of control'. A controller could then move the agent back towards its goal-directed pathway. This also points to the need for some way to formulate the goals, to state an *objective function* that embodies the goals. The rational agent behaves so as to create and execute whatever plan will maximize the expected value of the objective function. Can it be shown that a rational agent (in our case, a human agent) simply could not function without an explicit controller? These questions will be addressed next.

### 3.1 What Exactly is There to Control?

Vision, in its most general form, has been shown to have very bad computational properties; i.e., it requires too much compute power and too much compute time. This is a theoretical, many-times-proven, result with respect to a worst-case analysis of a general statement of the vision problem. There is no single, optimal algorithm that can provide a solution to any possible instance of any vision task within realizable compute and time resources. Evidence abounds for the intractable nature of even subsets of any generic vision definition, and some of these include: polyhedral scene line-labeling is NP-Complete[3] (Kirousis and Papadimitriou, 1988); loading shallow architectures is NP-Complete (neural network learning with finite depth networks) (Judd, 1988); relaxation procedures for constraint satisfaction networks are NP-Complete (Kasif, 1990); finding a single, valid interpretation of a scene with occlusion is NP-Complete (Cooper, 1998); Visual Match is NP-Complete (Tsotsos, 1989); stimulus-behavior search is NP-hard (Tsotsos 1995); and, 3D Sensor Planning for visual search is NP-hard (Ye and Tsotsos, 1996). Similar theoretical results based on computational complexity have also been presented by many researchers in artificial intelligence and cognitive science (too many to properly mention here, however, see van Rooij, 2008, for a nice review and van Rooij et al. 2019 for an extended treatment) and show how just about any non-trivial problem dealing with intelligent behavior is intractable in its general form. In practice, intractable here means that an optimal algorithm that applies to all problem instances requires time to compute that cannot be bounded by any polynomial function, and this is expressed as an exponential function the size of input, say $c^n$, where c is some constant and *n* is the size of the input (as a simplified example).

It is often difficult to imagine what an intractable problem instance, one with exponential characteristics as described, might look like. Imagine you are on a hilltop one evening with a crystal clear sky. The full array of stars above you stretch from horizon to horizon and you remember how as a kid you would look at the sky and wonder about the mythical constellations. Suppose you were asked to find a group of stars that formed a hexagon with equal length sides. You begin your search, and after a while realize you are checking every grouping of stars you can construct for overall configuration and side lengths. How many groupings are there of *N* stars? $2^N$, the powerset of *N*. How many groupings of 6 stars out of *N* stars? $N!/(6!(N-6)!)$. *N* does not need to be an unimaginably large number, but the number of stars visible in the night sky is pretty large. Even if it were only about a thousand, the number of potential groupings, $2^{1000} = 1.07 \times 10^{301}$, and the number of groupings of 6 stars is $1000!/(6!(1000-6)!)$ or about $5.0 \times 10^{134}$ are just too much to handle. You do not know what size the hexagon will be, what orientation it might have, where it can be found, nothing that you might use to restrict the group of stars that you consider. This is what an intractable instance looks like.

Does the natural visual world pose as difficult a problem as the constellation problem just described? Perhaps worse. Pavlidis (2009) concludes that it is impractical to construct training or testing sets of images that cover a substantial subset of all possible images. He estimates that $5^{36}$ is a lower bound on the number of discernible images ($5^{36}$ is about $1.5 \times 10^{25}$). Imagine the number of possible combinations of pixels - the powerset of the number of pixels - within such a large set of images. Recall how each object within a scene is a subset of pixels and a blind, brute-force search for potential objects needs to consider a great many potential regions. A dataset such as ImageNet with about $1.5 \times 10^8$ images really covers only 1 of every $10^{17}$ images possible. This number does not include video so these estimates are far too small. It seems completely impossible for every variation to be represented in the sufficient numbers of samples required to lead to good learning.

To be clear, the earlier statement that there is no single, optimal algorithm that can provide a solution to any possible vision task means that there is no single algorithm that can solve the problem for *any* value of *n*. van Rooij (2008) gives a nice summary where she lists problem classes that have such bad combinatorial properties that we can add to the list presented earlier. She provides formal definitions for each, including the general problems of Categorization, Visual Search, Gestalt Perception, Bayesian Reasoning, Planning and Network Learning. There is no reason to believe that any single methodology would change this basic reality nor is there any reason to believe that the brain can solve formally intractable problems.

---

[3] NP-Complete, in very simple terms relevant for the current discussion, refers to a class of computational problems for which all known solutions require super-polynomial time with respect to the input size.

The alternative is to re-shape the original problem into a different one that the brain can solve. As argued by Tsotsos (1990, 2011, 2017), such a problem re-shaping takes a particular form. The original problem can be addressed by partitioning the space of problem instances into sub-spaces where each might be solvable by a different method instead of having a single, optimal, algorithm for all problem instances (which is what the formal analyses just cited show is not possible). Why could a sub-problem be solvable when the full problem is intractable? Consider the problem of coloring a map of arbitrary regions. The constraint is that no two regions that share a border can be of the same color. It turns out that to determine *if* it is possible to do this with *k* colors, for any value of *k* and for any map configuration, is hard; but the same problem for *k=2* colors is easy. So a problem partitioning would be to develop one algorithm for the problem of 2 colors and others for $k > 2$ colors, rather than a single algorithm for $k > 0$ colors. This is an extremely simple example to illustrate the point, and there are many other examples. Problem instance partitioning is a tried-and-true computational approach to dealing with provably intractable problems (Garey & Johnson 1976).

It is also important to remember that even if problem difficulty is expressed as an exponential function of the size of its inputs, say $c^n$, given a powerful enough computer, it might be that some values of *n* can be handled thus making the problem partially solvable in practice. The resource constraints matter. This was the point of the *complexity level analysis* presented in Tsotsos (1988), to match problem difficulty to actual resource availability. The current success of learned deep networks is in part due to advances in compute power: the advent of GPU-level processors meant that they could handle sufficient size problems to become useful but they can not overcome the basic computational reality described above. The next topic to address is how might the full problem of vision be decomposed into more manageable pieces.

The kind of problem difficulty we describe here should not be confused with the notion of an ill-posed problem. If we look at the classic definition by Hadamard of a well-posed problem, mathematical models of physical phenomena should have the properties that: a solution exists, the solution is unique, the solution's behaviour changes continuously with the initial conditions. If a model lacks one of these properties it is termed ill-posed. Ever since the classic paper, Poggio, Torre & Koch (1985), visual information processing has been widely accepted a being an ill-posed problem. Often, it is easy to demonstrate multiple solutions (bi-stability) whereas the other conditions might be more difficult. Non-computational papers also acknowledge this and include Herzog, & Clarke, (2014) and, Nakayama & Shimojo (1992) and more. It might be thought that experience turns an ill-posed problem into a well-posed one (that is with a provable existence of a solution, with a unique solution that has continuous behavior with initial conditions), but this is unlikely. In Poggio et al. (1985), regularization was used as a tool to restrict the space of possible solutions and that if the space has finite dimensions, it might lead to a well-posed problem. However, as they note, not all problems can be framed so as to be amenable to regularization tools. On the other hand, they do not consider existence of a solution; this is addressed by a problem's formal decidability and detailed discussion of this for vision can be found in (Tsotsos 1993, 2017).

### 3.2 A Taxonomy of Visuospatial Problems

Human agents have a seemingly limitless capacity to develop solutions to the variety of visuospatial problems they confront. How particular visuospatial problems might be characterized has been long considered in psychophysical theory. Many of the particular problem names we use are borrowed mostly from their use in Macmillan and Creelman (2005), where one may look for definitions. As is likely intuitively obvious, the number of possible problems is not small so some form of organization is useful.

Some use an organization by number of experimental intervals (e.g., Macmillan & Creelman 2005, Luck & Vecera 2002). The term *interval* is used to signify the number of stimuli (as a leftover from auditory experiments where each sound occupied a separate time interval). Others use visuospatial factors that are exhibited when performing a task (Carroll 1993). These are not quite what we seek nor easily extended to the full breadth of visual capabilities. Here, we propose a scheme based on how many distinct retinal images are involved and their distribution in time. A visuospatial task will be termed an *n*-look task if it requires a subject to acquire *n* different retinal snapshots of the scene[4]. The period of time from onset of the task to its termination can also be encoded, specifying what time interval each snapshot covers and their order. All visual stimuli are included, e.g., blank screens, masks, fixation crosses, natural scenes. Examples will clarify: if a cue is presented on a screen and a subject is asked to note it, this is one observation; if a full scene is presented and the subject views it for 50ms, this is one observation; if the question to a subject is "is there a bird in this picture" and the image is presented for a short time interval with no possibility of eye movement, then one observation is made by a subject. These will be termed *1*-look tasks. If a sequence of *k* stimuli are shown one after the other with no opportunity for eye movements within any of those stimuli (such as in an RSVP presentation) then there is one observation per stimulus and the whole sequence represents a *k*-look task. All these tasks assume a passive observer; the subject performing the task is presented with stimuli and reacts to them. An active observer, by contrast, is one who purposefully chooses which stimulus to acquire for each observation, using either eye movements or full head and body movements to change viewpoints. A temporally constant stimulus may be part of an *n*-look task, *n >1*, if *n* eye movements are needed for its completion since *n* different retinal snapshots are acquired. A temporally varying stimulus would necessarily be part of an *n*-look task, one observation for each different retinal image. Different problem types can appear in different classes. For example, Visual Search can be a *1*-look or an *n*-look task depending of whether eye movements are permitted. Each of the problem classes in the taxonomy can be similarly considered depending on the temporal nature of the stimuli or goals.

Figure 2 shows a proposed problem decomposition. For clarity, the temporal dimension is not included. The point of this taxonomy is to show how it is possible to decompose the full, general, vision problem into smaller sub-problems that might help with the tractability issues described in the previous section. It is actually not important if this taxonomy is correct, unique or complete. It is important to connect parts of this to the complexity discussion earlier. That discussion proposed that the way to deal with a formally intractable problem is to split it up into more easily solvable subsets of instances. This taxonomy attempts to do this.

The Visual Match problem mentioned above is exactly the Recognition problem in the taxonomy: this is a decision problem (yes-no answer) asking if two stimuli are the same or different. The Sensor Planning problem is the same as any of the *n*-look group of tasks. The Categorization class is the problem addressed by the vast majority of CNN and Deep Learning vision systems. An orthogonal organization would be a compositional hierarchy (also known as a 'part of' hierarchy), that is, one that shows which problems are components of other problems. Of all the problems in the taxonomy, the most primitive class is Discrimination - distinguishing between two stimulus classes - in the sense of it being an element of all other vision problems. A part-of organization would have Discrimination as the root node that is part of all other problems. In other words, if Discrimination (or its more useful specialization Recognition, which is equivalent to Visual Match) is a component of all other vision tasks, and it is formally proved to be NP-Complete, then each vision problem includes an intractable component and is thus intractable itself.

Even though the theoretical conclusions described remain valid, the actual resource constraints for a human agent tell us that the human brain need not consider all possible sizes of inputs. Humans need only worry about those instances where images are retina-sized and not the full range of all possible image sizes. Tsotsos (1987, 1988, 1990) used this constraint to show how it is possible for the human visual system to deal with this size of image for simple vision problems (solvable by a single feedforward pass, that is, within 150ms). These vision problems have been well-studied (e.g., Potter 1975, Thorpe et al. 1996, Potter et al. 2014) with broad consensus on how simple categorization can complete within 150ms of processing time. As long as the targets of recognition are somewhat centered in the image (Tsotsos et al. 2019) and the image is not cluttered with interfering distractors (Yoo et al. 2019), this problem class within our taxonomy can be effectively solved by the brain. The impact of this ripples through the remainder of the hierarchy. As a result, there is no inherent paradox between the theoretical results and the reality of human vision.

---

[4] This is an approximation at best. It is acknowledged that there are many different retinal images taken even if a subject's gaze is perfectly controlled due to microsaccades (see Poletti & Rucci 2016). Further, visual input at the retina is analog and continuous rather than quantized in this manner. Nevertheless, this seems a useful organizational principle at this level of abstraction.

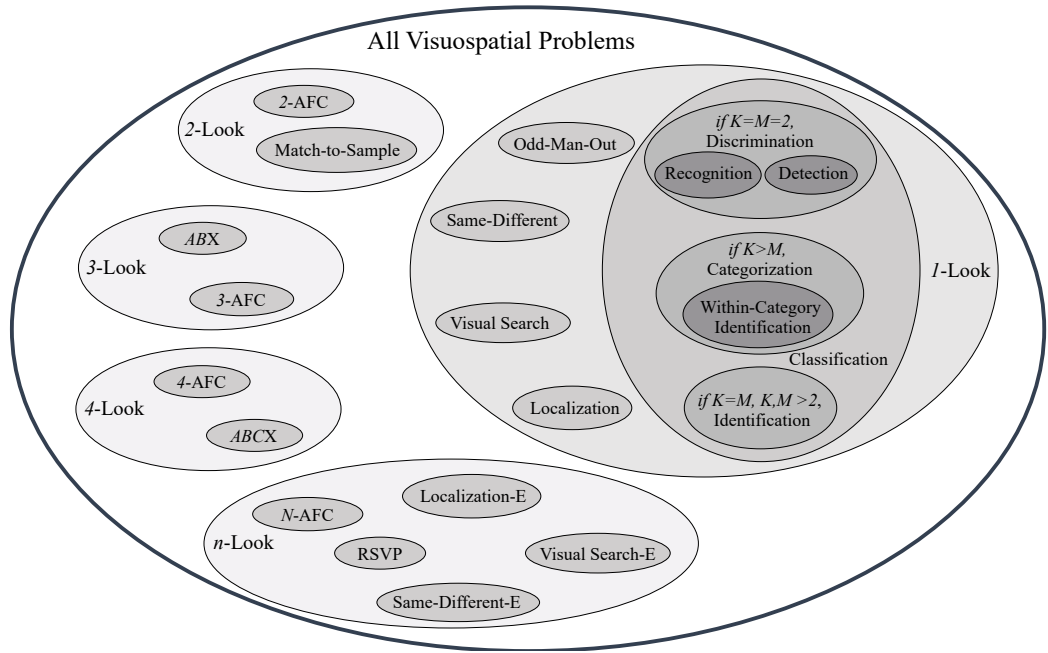

**Figure 2.** A suggested taxonomy of visuospatial problems. Each oval represents one class of the taxonomy and is labelled with the class name *(1*-look, *2*-look, *2*-AFC, etc.). The primary organizational dimension is the number of *looks* (separate retinal images) needed to successfully complete the task. The annotation "-E" signifies that eye movements are included. For example, one can perform a visual search task with a brief exposure and thus no eye movements (so it is a *1*-look task), or one can perform a visual task with free viewing (thus *n*-look). It is assumed that all of the other tasks are performed with fixed gaze; but any task can be performed with dynamic gaze and it is certain that there are many variations not in the taxonomy. Classification problems are sub-classified based on *K*, that represents the number of possible images, and *M*, the number of object categories of interest (following Macmillan & Creelman 2005). Descriptions for problem types can be found elsewhere (e.g., Macmillan & Creelman, 2005, Luck & Vecera, 2002, or similar).

Many other visuospatial tasks are easily apparent but are not included in the figure. Carroll (1993) presents a very thorough analysis of human behavior broadly speaking, devoting a chapter and more to behaviors involving visual perception. There, the factors that are used to group common abilities are suggested to be primarily[5]:

Visualization: apprehending, encoding, and mentally manipulating spatial forms (e.g., speeded tests such as Cards, Figures, and Flags);
Spatial Relations: compare two stimuli to determine whether one is only a rotated version of the other, or is a reflected (turned over) version of the other, rotated or not (e.g., Card Rotation, Cube Comparisons, Spatial Orientation);
Closure Speed: apprehending a spatial form that was in some way disguised or obscured by a noisy or distracting context. (e.g., Gestalt Completion, Concealed Words, and Mutilated Words);

---
[5] For more detail on the terminology used see Carroll (1993).

Closure Flexibility: searching a visual field to find a spatial form (specified in advance) despite a distracting context (e.g., Hidden Figures, Hidden Patterns, and Copying);

Perceptual Speed: either (1) searching a visual field for one or more specified spatial forms, without there being highly distracting or obscuring material, or (2) comparing two or more visual presentations for identity (e.g., Finding A's, Number Comparison, and Identical Pictures).

There are other suggested problem decompositions in the literature, but their consideration will not be helpful to our main point here. Carroll's volume emphasizes the following important points: there are several vision tasks Figure 2 does not include; alternate organizational dimensions are likely as valid as that of Figure 2; and, the number of such problems may be manageable from the perspective of building a taxonomy.

The previous section argued for a re-shaping of the vision problem. Vision should not be considered as a monolith because no single, optimal algorithm that solves all possible problem instances can exist. The taxonomy of Figure 2 proposes such a re-shaping. To complete the re-shaping, it is only required that all instances of each problem class be solved by the same algorithm. Just to be clear, an effective algorithm is a finite sequence of instructions to solve a class of problems or to perform a computation requiring tractable amounts of resources and time to complete. One algorithm may share elements with another, or may even include one or more other algorithms as sub-elements. However, overall, each solution differs (may require different resource requirements, have a different time course, and so on). The proposed decomposition does make one important demand: that all visuospatial behaviors can be represented by one or more elements of the taxonomy in combination (with appropriate parameterizations and computational 'glue'). For example, to solve a Visual Search task, one would require at least a sequence of Recognition (or Visual Match) tasks, each preceded by a selection of next fixation point. Finding some behavior that is not currently represented above would not falsify this decomposition; it could be added to the taxonomy or the taxonomy re-structured to accommodate.

The specific taxonomy is not as important as the act of specifying such a decomposition. It is critical to understanding how to provide solutions to each sub-task by limiting the scope that each proposed algorithm must address. It should be clear that there is no suggestion that the brain has such an explicit taxonomy. It might have some way of organization but the taxonomy here is purely for purposes of our analysis. However, it is suggested that the brain's solutions to each of these tasks differs; that is how the brain overcomes the theoretical intractability described in the previous section, by not relying on a single algorithm for all problem instances.

Our rational agent now needs one more element: a method that, when confronted with a visual problem instance whose solution has resource constraints, can quickly determine which solution method to apply. In other words, in our everyday life, how do we invoke a particular problem solution at the right time and in the right context? Our visual systems are not simply machines that decide the category of the foreground object in pictures shown to them. Our brain needs to determine which kind of solution is needed at any moment in time depending on the stimulus and our purpose in viewing it. Solutions may involve one of the problems in the taxonomy or a sequence of problems assembled dynamically for a particular goal of the agent at a particular time and in a particular context. In essence, what this discussion accomplishes is to argue for why a *Task Executive Controller* is necessary (one of STAR's basic 6 components). An organization of the problem space acts as a tractable foundation for the operation of such a controller.

## 3.3 Is Attentional Control Necessary in Vision?

We examine five possible answers to this question (the last will appear in the next section). First, it may be that attentional control is not necessary at all, that any attentional effects we observe externally simply emerge. The Gestalt school did not believe in attention; Köhler only barely mentions attention (Köhler, 1947). Gestaltists believed that the patterns of electrochemical activity in the brain are able to sort things out by themselves and to achieve an organization that best represents the visual world, reconciling any conflicts along the way. The resulting internal organization includes portions that seem more prominent than others. Attention, to them, was an emergent property and not a process in its own right, and then it would go without saying that attention does not need a controller. Figure-ground concerns loomed larger for them, the figure would dominate perceptions within a scene, thus emerging as the focus of attention rather than being explicitly computed as such. Duncan (1979) provided an early discussion of properties of attention having an emergent quality in the context of divided attention. Pomerantz and Pristach (1989) and Treisman and Paterson (1984) worked on emergent features in attention. Desimone and Duncan (1995) seem to suggest that their biased competition model is emergent, writing "attention is an emergent property of slow, competitive interactions that work in parallel across the visual field.". This is essentially the same argument behaviorists have been making since the early 1900's, and strengthened by the success of Rod Brooks' robotics program in his early years at MIT (e.g., Brooks 1991). There are two points here. First, how many behaviors are we talking about? A bee, for example, does not have a large repertoire; humans have a seemingly infinite one. So what one sees as emergence for bees may not translate easily to humans[6]. Second, perceptual abilities - and it's a percept that triggers a behavior - are also limited in bees while again seemingly infinite in humans. Both of these make behaviorism or emergence as the answer for the totality of human visual behavior implausible (see Tsotsos (1995) for proofs and quantitative arguments; an earlier paper, Kirsh (1991) provides additional logic and philosophical arguments).

The key point remains that humans have an ability to correct failing behaviors and this requires more than simple emergence. The problem here is that any process from which behaviors might emerge would not also be able to track that behavior's performance, to detect when it fails, nor to then reason about how to repair or refine the behavior so that it can succeed. The reason follows directly from the nature of emergence: an emergent behavior is the result of some set of sub-behaviors whose sum of effects (and side-effects) provides the desired behavior. Thus, there is no element of the overall system that would know in advance what the behavior should be as would be necessary for solving any specific task, nor would it know how to adjust those elements if a particular arrangement fails to produce the desired behavior (see Steels 1990 for more on emergent behavior). In an emergent system, there is little or no central control whatsoever; neurons and brain areas operate in an uncoordinated, asynchronous manner. One would also need to understand how the precision observed in human behavior comes about (think about the precision required to properly return a tennis serve, to judge exactly how deeply and with what force a dentist drills your cavity, etc.). Emergence seems inadequate as a substitute for explicit control.

A second answer is that there is no need for feedback in vision and thus no top-down visual attention, that it can all proceed in a purely feedforward manner. Any task or knowledge-specific influences are reserved for higher levels of cognition, but not within vision, as Marr (1982) advocated. A very nice argument to refute this is provided by Herzog & Clarke (2014) who review a wealth of evidence suggesting that cortical processing can not be purely hierarchical as well as purely feed-forward. The visual system processes fine-grained information at a particular location by integrating information about the surrounding context over the entire visual field. Grouping and segmentation are crucial to understanding vision, and must be understood on a global scale, they point out. This cannot occur without top-down communication (Tsotsos 2011, 1990).

---

[6] It may be useful here to include a quote from Brooks (1991): "Like Minsky (1987) we believe that human level coherence during many activities may only be in the eye of the beholder; the behavior is generated by a large collection of simpler behaviors which do not have the rationality generating them that we might normally attribute to humans."

A third possibility is that there is indeed control but that it is a passive, fixed, hard-wired process, that operates kind of like a decision-tree that is traced for each task in order to find the best control actions. This would mean that there are sufficient separate hard-wired instantiations to account for the seemingly infinite flexibility our visual systems seem to have. An example may be useful at this point to illustrate the difficulty of hard-wiring all visuospatial behavior.

The Same-Different task is one of those Carroll gives in the category of Spatial Relations. Its classic expression is to show a subject 2 three-dimensional shapes on a two-dimensional screen or piece of paper. and ask whether the objects are the same or different. The objects may be identical but rotated so different views are seen, or similar but different and rotated so as to make their differences non-obvious. Observation in this sense is passive since a subject sees only a two-dimensional projection of the objects with no possibility of changing viewpoint (so as to reduce self-occlusions, for example). We have examined an extension of this problem, in a 3D active observer setting. The two stimuli to be compared are three-dimensional real objects, sometimes identical but rotated in 3D and sometimes different and again at random rotations, and the observer is free to move about and view the stimuli from any viewpoint that might be useful to the task (i.e., the observer is an active participant). Some observed action sequences are shown in Figure 3 (from Tsotsos et al. 2018).

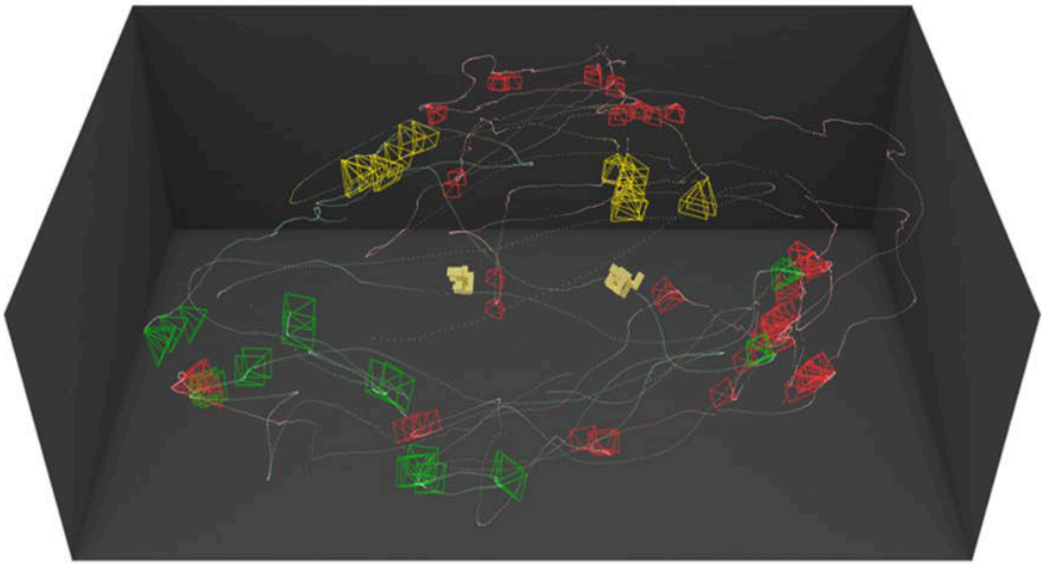

**Figure 3.** A visualization of trials performed by three different subjects (red—R, green—G, yellow—Y) under the same conditions. Two different objects were placed in the middle of the room, with poses differing by 90°. The trajectory of each subject is plotted in the corresponding colour, and each observation of the objects is displayed with an oriented frustum. (from Tsotsos et al. 2018).

This example suggests that there seems to be no simple characterization of this behavior and in fact, our current analysis of 66 subjects performing a total of about 1200 trials leading to 11 million tracking points concludes that subjects exhibit a set of many different strategies, each chosen depending on the stimulus characteristics and on the starting viewpoint in a trial (Solbach & Tsotsos 2020a; Solbach & Tsotsos 2020b). In other words, even a single class within the taxonomy of Figure 2 really needs more elaboration. That is, the Same-Different class involves a variety of different solution strategies or algorithms, each appropriate for a particular sub-group of problem instances. Moreover, as is clear from the traces of head-body motion in the figure, subjects choose what viewpoint to use at any point in their trial, and decide when to change fixation from one point to another in a purposeful manner since they do eventually solve the task with high accuracy. It would be very difficult to argue that no attention is needed in order to do this: some method to choose what to attend to and in which order is required.

Attention, like vision, is not a monolithic entity. There are many well-known manifestations of attention of which only some are mentioned for the Same-Different task. Adapted from Tsotsos (2011), the following may be considered as the set of attentional mechanisms observed in human vision (citations to justify each can be found in Tsotsos 2011):

> **Alerting** The ability to process, identify, and move attention to priority signals.
> **Binding** The process that correctly combines visual features to provide a unified representation of an object.
> **Covert Attention** Attention to a stimulus in the visual field without eye movements.
> **Disengage Attention** The generation of signals that release attention from one focus and prepare for a shift.
> **Endogenous Influence** Endogenous influence is an internally generated signal used for directing attention, including domain knowledge or task instructions.
> **Engage Attention** The actions needed to fixate a stimulus whether covertly or overtly.
> **Exogenous Influences** Exogenous influence is due to an external stimulus and contributes to control of gaze direction in a reflexive manner.
> **Inhibition of Return** A bias against returning attention to previously attended location or object.
> **Localization** An item is attended if it is selected and then all its defining characteristics are identified. This requires a top-down process.
> **Neural Modulation** Attention changes baseline firing rates and firing patterns of neurons for attended stimuli.
> **Overt Attention** Also known as **Orienting**, the action of orienting the body, head, and eyes to foveate a stimulus in the 3D world.
> **Priming** Priming is the general process by which task instructions or world knowledge prepares the visual system for input. **Cueing** is an instance of priming; perception is speeded with a correct cue, whether by location, feature, or complete stimulus. Purposefully ignoring is termed **Negative Priming**. If one ignores a stimulus, processing of that ignored stimulus shortly afterwards is impaired.
> **Salience/Conspicuity** The overall contrast of the stimulus at a particular location with respect to its surround.
> **Search** The process that scans the candidate stimuli for detection or other tasks among the many possible locations and features in cluttered scenes.
> **Selection** The process of choosing one element of the stimulus over the remainder. Selection can be over locations, over features, for objects, over time, and for behavioral responses, or even combinations.
> **Shift Attention** The actions involved in moving an attentional fixation from its current to its new point of fixation.
> **Surround Suppression** Local context is suppressed via top-down branch-and-bound process to reduce interference to an attended stimulus.

It is easy to disagree with this set of mechanisms, to think that this set is not complete, that one or another element should or should not be present, and so on. Moreover, there are alternate views of this set. For example, Gilbert & Li (2013) suggest the set of top-down attentional influences include: spatial attention, object-oriented attention, feature-based attention, perceptual tasks, object expectation, efference copy, working memory, associative memory and perceptual learning. These are overlapping with our own list, the levels of abstraction are not the same (e.g., 'shift attention' is really an implementation element of all their mechanisms), but still they provide 9 different roles. However, it is easy to agree with the fact that there are several manifestations of attention, each seemingly due to separate mechanisms. The literature is large with their descriptions (see Itti, Rees & Tsotsos 2005, Carrasco 2011, Nobre & Kastner 2014).

Returning to the Same-Different task, we could imagine a Cognitive Program for one possible strategy. One of the observed strategies involves selecting one feature on one object and then moving to view its hypothesized counterpart on the other object to check to see if it is present. This repeats until sufficient features are confirmed to signal a 'same' response or until a difference is found which would lead to a 'different' response. This solution involves sub-sequences of select fixation on one object, move fixation, engage attention, extract features, store in working memory, disengage attention, hypothesize pose of candidate features on other object, select new viewpoint for candidate features, move to new viewpoint, select fixation point, move to fixate, engage attention, extract features, match to contents of working memory, decide if successful, store decision and completed feature matching in working memory, give final response, or repeat. This illustrates the connection between the problem taxonomy and Cognitive Programs. Each CP is a composition of algorithms that solve one or more problems in the taxonomy, parameterized for the current task and stimulus, and linked by communication and decision-making computational glue.

Even for the most basic of the visual tasks of Figure 2, say Recognition ($K=M=2$), the sequence of events is non-trivial. Suppose a subject is presented with a trial in an experiment whose goal is to examine recognition performance. If there are two classes from which to choose membership of a stimulus, the subject needs to first be presented with those choices. Presentation might be through the name, a picture on a screen or through sound. In any case, a representation sufficient for the task needs to be extracted from the cue and stored in working memory. Next, it might be the case that a fixation cross is presented with instructions to not move one's eyes away from it. The visual system must be instructed to do so, then it must shift its attention, engage it spatially at the fixation, and suppress subsequent eye movements. So far, a number of the attentional elements listed above must come into play, all before the actual stimulus test. Then the stimulus appears, it must be processed and visual characteristics extracted for comparison with what was in working memory (again extracted by the same visual system for the cue presented visually), and a comparison decision made, followed by a response. It might be that this sequence can be learned, and then applied as needed. But to learn it in the first place would involve the composition of the correct elements; and then for any new scenario a new composition would be needed, and so on. The flexible, combinatorial and hypothesize-and-test nature of how a subject learns to perform a visual task should be clear.

Let us say that *A* is a variable that represents the number of such attentional mechanisms and that its exact number is not important, only that it is a non-trivial number. Given this, the next question is for a given task, which attentional mechanisms are needed? Secondary questions include: in which order should they be deployed, how should each one operate, what is the best parameterization for each, how will it determined that they succeed, how will they be re-defined if they fail, and how will they be reset for the next task? We address only the first question here. In a system that has no prior guidance to draw upon, all possible subsets of mechanisms would need to be considered in order to choose the best one (recall the constellation example). This means there are $2^A$ such subsets; and if we consider their possible parameterizations, their durations and the fact that they might have repeated instances within a solution sequence, the exponent goes from *A* to a much, much larger number. Even given the set of mechanisms above with $A = 19$, this means $2^{19} = 524,288$ possible subsets. Even if each has only 5 possible durations ($d = 5$), 2 different parameterizations ($p = 2$), and none are repeated, $2^{Adp} = 2^{190} = 1.6 \times 10^{57}$. This gets out of hand and out of the realm of possibility very quickly; *d* may take on many more values than 5, the same is true for *p,* and if repeated actions are permitted, e.g., the need to attend to two different items at different times, this is completely impossible to deal with.

Even if it were possible to have all these possibilities hard-wired (the number of separate combinations of neurons, synapses and connecting pathways is extremely large), one would still need to select a stimulus sample out of the environment, note what that environment might be, select which set of hardwired paths are needed, initiate, monitor and correct if needed, and finally decide on an output behavior. This is a very difficult possibility to accept. Of course, one may argue that although this number is very large, not all of these are useful combinations. Only a small number might be ever used in a lifetime so there is no point to address the full problem; it suffices to find good solutions to the set of commonly seen or 'expected' situations. Such an argument was made by Lowe (1990) against a very similar conclusion made in Tsotsos (1990). Lowe said that the combinatorics argument was not useful because the visual system evolves to handle the situations commonly encountered and not all possible ones. The analysis Lowe was critiquing was the usual worst case analysis performed for computational complexity proofs. In other words, the analysis in Tsotsos (1990) was based on a worst-case analysis, one where problem instances of the greatest possible difficulty were part of the overall consideration. Lowe's point was well-made, and led to a detailed empirical investigation of the difference between median case (as an approximation to expected case) and worst case analysis. The result was that even for the median case, solutions still had exponential characteristics (Parodi et al. 1998). Only when task knowledge (very small information about search order) was added did that performance become linear (as was also shown in Tsotsos 1989 for the Visual Match problem; it has linear time complexity characteristics when guided by task information). This is a very strong evidence that top-down knowledge plays a very important role in combatting the inherent combinatorics of visual processing (it would have, for example, helped you in the constellation example to know the expected size of the hexagon since that would limit the search considerably).

A fourth option is that there is sufficient computational power in the brain via parallelism to allow some kind of brute-force search to succeed in rapidly determining the appropriate set of attentional mechanisms needed for a given task. In other words, the possibilities could be encoded in advance somehow and enough computational power is available to very quickly search through them on demand, testing each to see if it were the right one. Estimates of the number of neurons in the brain is 86 billion (Azevedo et al. 2009) with the number of synapses perhaps 10,000 per neuron on average. Suppose that each possible set of mechanisms could be decided by a single synapse. This would mean that the number of processors acting in parallel to process the number of potential alternatives would be limited to $8.6 \times 10^{14}$. This would permit a value of *Adp* of at most 50 ($2^{49.62}= 8.65 \times 10^{14}$). In other words, for our set of 19 attentional mechanisms, the variation for each could be only about 2-3 different parameterizations. This seems wildly inadequate. Allowing each synapse to examine possibilities in sequence would not help. This of course would mean the entire brain does nothing else for that time interval; this is not the case. The brute-force strategy just does not seem right.

The first two possibilities dealt with scenarios where there was no control at all while the next two dealt with possibilities that a complete controller could be implemented to cover all vision tasks in a passive manner (no 'thinking' involved). Arguments that all are unlikely have been presented. This leaves only the possibility that some kind of active controller (that 'thinks') is what is actually present in the brain. Thus, the fifth possible answer to our original question is that there is a flexible control process that actively deploys the needed mechanisms for the task and environment at hand in a dynamic fashion. This is consistent with the conclusion and main claim of Tsotsos (2011), that this kind of dynamic attentional process is necessary because the machinery of the brain is far too small to permit the flexibility and generality of function of human visuospatial behavior. In order to do this, there must be some method that overcomes the combinatorics described above. This is the topic of the next section.

# 4.0 A Sketch of an Active Attention Controller

At the beginning of Section 3.0 we asked two questions: First, what exactly is there to control? Would a human agent simply not function without an explicit controller? For the first question, we reviewed extensive research that has shown that the general problem of vision is too difficult for any single optimal algorithm solve. The solution is to *divide-and-conquer*, to identify sub-problems each with different algorithms for their solution. Some of those algorithms will be efficient, some perhaps less so, but this means that there is a sub-space of the overall problem which does have good solution. We presented a taxonomy of vision problems and each sub-problem would have its own algorithm, resource requirements, and performance characteristics. This means that some method to decide which solution to deploy at which time is needed, i.e., a controller. For the second question, we described how the algorithms that solve each of the visual problems in our taxonomy have potentially several attentional elements. These attentional mechanisms could not appear due to emergent behavior, brute-force computation through all possibilities nor any hard-wired circuits. Their synchronization, parameterization and coordination requires a controller. A sketch of such a controller is now outlined that might be able to defeat the combinatorics of the problem.

In Tsotsos (1987, 1990) the intractability of the general vision problem led to a re-shaping of the problem. Key elements of the re-shaping included a reduction in the search space using physical constraints of the visual system, the use of task specific top-down processes, and a re-organization of the processing architecture to take advantage of the performance improvements afforded by hierarchical structures. Certain aspects of the general problem were lost in this way, specifically the major one being that only the spatiotemporally localized receptive fields were included and not all possible combinations of input elements. In practice this meant trading off those combinations for processing time. We can consider similar tools for our attentional controller.

Many readers might naturally wonder why in this age of deep learning our approach for attentional control would not take advantage of its power. More discussion around this question appears later, but for the moment, the following brief points are mentioned. A learning approach assumes that a sufficient corpus of the relevant data for training can be obtained. It also assumes that the 'answer' to some question actually lies within the statistical characteristics of the data. One would need to know both the question and what an answer might look like in advance; neither are clear here. Finally, it seems that both past experimental work as well as deep learning approaches in any domain neglect important questions: how is progress towards a problem's solution monitored during its execution, how are deviations detected, and how are are such unplanned failures corrected? It seems clear that in order to monitor the progress of an algorithm's execution, the ability to inspect some or all intermediate stages and representations is necessary. The intermediate layers of representation found in deep learning systems do not easily lend themselves to this. Such a blanket statement is easily challenged; there is some evidence on inspection of neurons in early layers and many are active in trying to better understand the representations. However, it is not at all straightforward as to how to use the distributed representations that pervade network architectures in the localized fashion the answer to this question seems to require. This is not to say that this can never be solved; perhaps it will, but for now, such representations do not seem to have the right character for the issues of plan monitoring and repair. We thus continue exploring our hypothesis.

As was discussed earlier, the use of task-specific guidance can turn an exponentially difficult problem into a linear one. To apply this idea to how an attention controller might defeat the combinatorial nature of the set of all possible combinations of attentional mechanisms means that any search through that enormous space of possibilities can be restricted to only those relevant to the task at hand - on the assumption that some task is actually available. The case where this assumption is not valid will be discussed once the positive case is addressed. If the task is, for example, Recognition (recall the 'do you see your mother's face' example), only a limited number of the full set of attentional mechanisms need be considered and the impact is an exponentially smaller amount of computation. An illustration of the magnitude of improvement follows.

In demonstrating the combinatorics earlier, the number of attentional mechanisms was shown in the exponent of the expression; task knowledge can make exponentially important reductions to the search space. Consider our constellation example again. Suppose that you know that the hexagon in the sky that you seek has a size of 20° of visual arc (out of the 180° that the full sky covers along each horizon to horizon arc.) How does this change the calculation? We need to compute the value of $N!/(6!(N-6)!)$ where N still is 1000 but where the acceptable groupings of 6 have the characteristic that no distance between any two stars is greater than 20°. For the sake of the back-of-the-envelope calculation, assume that the stars are uniformly spread over the sky. The sky can be approximated by a half sphere whose surface area is given by 62831 units (surface area of a whole sphere is *$4\pi r^2$* and we can assume a sphere with *r=100*). There would thus be about one star per 63 surface area units. A 20° x 20° visual arc surface patch has an area of 1193.9 surface units, i.e., there are 52.6 such patches on the half-sphere, and thus each contains 19 stars. The search could then involve checking the groupings of 6 stars centered at 62831 such discs (one centered at each unit of surface area; this incudes overlap to ensure all star groupings are counted). This means $62831 \times 19!/(6!(19-6)!) = 1.95 \times 10^7$ possibilities, a 127 orders of magnitude improvement from the earlier calculation where there was no task information. Actually, this over-counts; at the horizon full discs would overflow below the horizon, so the actual count is smaller still. To take advantage of task guidance requires that the overall process be adaptable and flexible to such knowledge, i.e., be an active one whose parameters are set dynamically, and the advantages are very clear.

If a specific task is not available - if you are taking a walk on a warm sunny day in a new town just taking in the sights and sounds - then what? What can be perceptually extracted from the stimulus can play the same search limiting role, although in a different manner, with a potentially different result and with a different time course. Recall the classic cueing experiments from Posner et al. (1978) and so many more. They clearly show that if a cue provided in advance of a stimulus is valid, performance is both faster more accurate. If the cue is invalid, performance is poorer and slower. If there is no cue, performance falls somewhere in between these two. While walking down the street, your visual system takes in the scene. It is processed normally, without a specific task (although it seems impossible to eliminate any sort of contextual influence or your own past experience in such situations) and after the first feedforward pass though the visual system, certain concepts are noted with higher confidence than others. These then could provide cues for subsequent processing. The appearance of any such guidance might be a bit delayed in comparison to a specific task, and it might lead to different sorts of perceptions, but these play a similar limiting role on the search space of mechanisms. This has more than a passing similarity to what von Helmholtz (1867) pointed out, that perception proceeds by testing experienced-derived hypotheses against sensations.

Even with such a dramatic reduction in the search space, this still seems a computationally demanding task especially if one considers the huge breadth of visual behaviors a normal adult human commands. However, humans are not born with the full capacity to solve all possible vision problems. Siu & Murphy (2018) show a comprehensive survey of development where they highlight the changes in the visual system from birth onwards. Full visual function is not available until late adolescence. We have looked at a specific attentional function, the attentive suppressive surround, and confirmed such a late development (Wong-Kee-You et al. 2019). We found that this suppression is not seen in children at all until about 7-8 years old and only matures by 17-18 years old. Siu & Murphy further point out that feedback input within the visual hierarchy is not available fully until 2 yrs, spatial acuity and contrast sensitivity are mature at 8 yrs, face perception is not fully mature until 18, and even feedforward input is not mature until 6 months. Much of this delay is due to the maturing of physiologic and anatomic substrates needed for these functions. This tells us that the brain does not address the full combinatorics of vision at once; it does so in a gradual manner that takes many years to develop[7]. This means that our attention controller could have a starting point that is very basic in its functionality and gradually matures over time. Experience with visual behaviors over time then adds an additional filter to the large set of possible attentional mechanisms enhancing that of task.

What could the starting point be? Macmillan & Creelman (2005) give us a clue. They write "The basic psychophysical process, we believe, is comparison. All psychophysical judgements are of one stimulus relative to another, designs differ in the nature and difficulty of the comparisons to be made." They go on to highlight that the matching procedure is critical to comparison: "To judge the subjective magnitude of a stimulus, a participant selects a value on some other continuum that seems to 'match' the standard. For example, the brightness of a light might be matched by the intensity of white noise or the brightness of a light of a different color." Suppose that our attention controller starts with the ability to perform comparison; this is essentially the ability to perform the Discrimination task of our taxonomy in Figure 2. As the brain matures, it works out the details for Discrimination (parameters, timing, etc.) and perhaps extends it to the next problems in the taxonomy (e.g., Recognition, Classification). As the physiologic and anatomic substrates mature, next problems can be addressed based on these early ones (e.g., problems requiring eye movements for example since according to Siu & Murphy binocular fusion and stereopsis are mature by 6 months). If a reduction in search space for developmental reasons (i.e., the exponent *A* has a very small initial value) is combined with a reduction due to task characteristics, the overall search space could indeed be a very small and thus easily manageable particularly in early years of development (likely the developmental reduction is greater in the earliest period with task playing a role once the notion of task can be understood later). It is clear that this is a hypothesis at this time from the human point of view; from the computational point of view, however, the reasoning is sound and the four elements proposed as the solution for overcoming the combinatorial issues will certainly suffice. They are: a) re-shaping the problem space, associating each problem class with an appropriate algorithmic solution; b) active task influence to reduce search space; c) experiential learning to reduce the parameter search space; and, d) relying on a controller to orchestrate. It seems likely that all of these elements are necessary, but this proposition requires additional examination.

## 5.0 Computational Goals of Attentional Control

Our target problem is attentional control in vision. We have addressed the 'attention' and the 'vision' pieces to some degree in the previous sections, but not yet the 'control' element. As mentioned in Section 2.0, an accepted way to formulate the goals of a system from a control standpoint is via an *objective function* that embodies the goals. The system then behaves so as to create and execute whatever plan will maximize the expected value of the objective function.

It is premature to think that the full executive control problem might be encapsulated within a single objective function (but see Friston 2010) and it would be beyond the scope of this paper to make the attempt. However, we will attempt to specify the highest level of description needed for Marr's computational level - What is the goal of the computation? - by hypothesizing the objective function for an attentional controller.

---

[7] With all respect to A.M. Turing, his assertion, that seems to anchor modern machine learning in developing computational agents with human-like intelligence, can no longer be considered as valid. He wrote (Turing 1950): *Instead of trying to produce a programme to simulate the adult mind, why not rather try to produce one which simulates the child's? If this were then subjected to an appropriate course of education one would obtain the adult brain. Presumably the child-brain is something like a note-book as one buys it from the stationers. Rather little mechanism, and lots of blank sheets. (Mechanism and writing are from our point of view almost synonymous.) Our hope is that there is so little mechanism in the child-brain that something like it can be easily programmed. The amount of work in the education we can assume, as a first approximation, to be much the same as for the human child.* Certainly, in his day, Turing made the best of what was known; but in 70 years this has changed.

An optimization problem is typically represented as a function $f: A \to \mathbb{R}$ from a set $A$ to the real numbers and a solution $\mathbf{x_0} \in A$ is the goal such that $f(\mathbf{x_0}) \leq f(\mathbf{x})$ for all $\mathbf{x} \in A$ (this is for a minimization goal) or such that $f(\mathbf{x_0}) \geq f(\mathbf{x})$ for all $\mathbf{x} \in A$ (for a maximization goal). The set $A$ represents the search space of candidate solutions and the domain might be specified by some set of constraints to delineate the acceptable subset of the Euclidean space $\mathbb{R}^n$. Each candidate solution is represented by a vector $\mathbf{x}$ of all the relevant variables that are needed to define the solution. The function $f$ is referred to in several ways: as an objective function, as a loss of cost function if minimized, as a utility or fitness function is maximized, and as an energy function. A feasible solution that satisfies the objective function (whether minimization or maximization) is called an optimal solution. Although a good objective function provides a formal statement of the goal for problem solution, it would not be useful without some means for the system to which it applies to make changes that attempt to achieve the objective. As mentioned above, a rational agent behaves so as to create and execute whatever plan will maximize the expected value of the objective function. This means that what ever system component examines the status of the objective function must also provide the appropriate signals in order to move the overall system closer to its goal. We will term these the *Type I control signals* and there would be at least one control signal for each variable relevant to the objective function (each variable of the vector $\mathbf{x}$) that would push overall processing in the right direction. *Type II control signals*, on the other hand, are signals that coordinate timing of processes, e.g., to initiate and terminate processes. There may be other types of signals as well. Each attentional mechanism may have more than one control signal relevant to its operation (e.g., *disengage attention* might require only an initiate signal while *IOR* may require an initiation signal, a decay rate parameter, a location, and so on). Finally, each process would also pass data to other processes (e.g., the result of an image convolution); likely there may be other sorts of information that process communication might involve.

Expressing a problem as optimization (rather than decision with a yes-no solution) does not change the inherent computational difficulty of the problem; any inherently intractable problem remains so (Krentel, 1986; Papadimitriou 1993). Common optimization problems that are difficult, and form the foundation of most current AI methods that employ optimization, include integer linear programming, travelling salesperson, coloring, clique, knapsack (Krentel 1986).

It is interesting to note one particular element of such a control formalization. It depends on *comparison* to determine whether the function is satisfied or not and for the direction and amount in which to affect change. This connects nicely to the role that Macmillan & Creelman (2005) suggest for comparison as the basic psychophysical process, described earlier. Similarly, optimization frameworks require some method of measuring the deviation between observed and reference (or desired) variables. We can specify this deviation as $\Delta_i(K_r^i(t), K_\gamma^i(t))$, where $\Delta_i$ is some appropriate measure for the $i$-th control variable. There can be many different ways of measuring such differences, each specific to a particular controlled variable. $K_r^i(t)$ is the reference value of the $i$-th control variable (counterpart to the elements of $\mathbf{x_0}$ at time $t$. $K_\gamma^i(t)$ is the observed value of the $i$-th control variable under the control signal $\gamma(i, t)$. Most common objective functions specify the comparison as a subtraction between real-value numbers. However, this seems too restrictive for our goals.

We are now ready to state the control problem at Marr's computational level. The controller seeks to maximize the rationality of the agent's behavior. In other words, it seeks to minimize any 'out of control' activity, recalling the statement made in Section 3.0. The global control objective is to seek the set of control signals $\Gamma=\{\gamma(1, t), \gamma(2, t)..... \gamma(N, t)\}$, for $N$ control processes at time $t$, that satisfies:

$$\min_{\Gamma} \sum_i^N \Delta_i(K_r^i(t), K_\gamma^i(t))$$

$N$ thus represents the set of all Type I and Type II control signals. For the Type II signals, suppose that a particular visuospatial problem is required to complete by time $t_c$; it is initiated at time $t_i$. The CP sets the relevant values of $K_r^i(t)$ to be 0, for $t < t_i$ and $t > t_c$. Any individual CP would involve some subset of these $N$ signals. It is important to note that this set of control signals cannot be fully determined in advance in any real-world setting. There may be certainly default values but in general, the defaults would be modified as execution of the CP progresses to adapt to the task, environment and system performance in that environment. We now consider some examples of such signals and objective functions for a few CPs.

We consider three specific examples, the first dealing with eye movements. The point of an eye movement is to bring the retina's region of highest acuity to the current location of interest in a scene (Tsotsos et al. 2014). In the specification of eye fixation change it might be useful to give the goal as follows. Let $S$ be the hypothesized image element of interest, selected in advance. Since the fovea, and specifically its foveola, has the highest density of cones, it should always be placed at the location of maximum interest, or if there is no particular location of interest, then at the centroid of $S$ as a default. Let the gaze position be $(x, y)$ in retinal coordinates and the centroid of $S$ be $(x_S, y_S)$. The closer the point of gaze - or the center of the foveola - is to the object centroid, the more retinal cones will fall within the object, so a controller will seek to minimize the distance between these two points with the following objective,

$$\min_{(x,y)} \|(x_S, y_S) - (x, y)\|$$

at any time $t$. Here, $\Delta_i$ is the $L_2$ norm (Euclidean distance), the reference $K_r^i(t)$ is $(x_S, y_S)$ while the observed value $K_\gamma^i(t)$ is $(x, y)$. The control signal for the fixation mechanism would be $(\delta x, \delta y)$, the change from the current $(x, y)$.

A second example deals with the problem of feedforward signal interference in a hierarchical network, of which there at least 2 kinds, one due to feedforward diverging connectivity patterns (one neuron to many) and the other due to feedforward converging connectivity patterns (many neurons to one). We term the first the Crosstalk Problem and the second the Context Problem (Tsotsos et al. 1995, Tsotsos 2011, 2017). Each of these problems impact every neuron in a layered, hierarchical network; they are inherent in the architecture. The result is that although the visual hierarchy can see (encode) everything in its input, it might not always be able to distinguish one thing from another because of the interference (see effective demonstrations in Rosenfeld et al. 2018). Such interference is common in other domains and a well-known solution is available, namely, adaptive beamforming (Tsotsos 2017). An adaptive beamformer is a system that performs signal processing by dynamically manipulating the combination of signals (how they interact and interfere with each other) so that the signal strength to/from a chosen direction is enhanced while those to/from other directions are are degraded (this is commonly used in cellular communication). Adaptive beamforming seeks to maximize the signal-to-interference-plus-noise ratio $S/(I+N)$ (Vorobyov 2014). In order to use this method – dynamic control of constructive and destructive interference – a localized representation is required. A distributed representation (such as one where all concepts are represented using a code only recognizable by considering a population of responses) could not be so controlled.

Generally, $S$, $I$ and $N$ are not known in advance. The interference $I$ is dependent on context but defined by network connectivity (e.g., context is whatever is in an image that is not the target of expectation, the room background in the example given above). The noise $N$ is dependent on context (the stimulus and the method of its sensing has its own sources of noise) and on neural processes and their inherent stochasticity. Noise cannot be controlled, but $I$ can be controlled because the way information flows through the processing network can be manipulated.

The signal of interest $S$ is dependent on input and goals and can only be hypothesized or predicted (e.g., expect to see your mother's face). If one has a predicted $S$ a priori, then it is used to prime the hierarchy in advance of the stimulus; however, in many situations there is no a priori expectation. Here is where the details of how ST implements the process provide a solution. When there are no expectations, the input stimulus is processed in a generic fashion - all interpretations are equally likely. Naturally, the actual stimulus will flow through the hierarchy on its feedforward pass and by the time it reaches the higher levels, some concepts will exhibit higher levels of response. ST would then choose the best of these and hypothesize that it represents *S*, and use it to suppress its hypothesized interference. To suppress any hypothesized interference, it implements a hierarchical, top-down, branch-and-bound process[8]. This is inherently a search process that makes local decisions during a top-down traversal of the visual hierarchy, keeping nodes consistent with the expectation while suppressing or discarding the others. Sometimes, decisions are made that may not be the correct ones because of noise or incorrect predictions. Thus, there needs to be some way to detect incorrect decisions. Neurons chosen at the top are hypothesized to be most representative of the required focus of attention, and if correct, the irrelevant neurons at that level are suppressed[9]. This means that the response of the selected neurons will monotonically increase because the effect of the irrelevant is reduced, as observed by Reynolds & Desimone (1999). Let the response at time *t* of the neuron that represents *S* in network layer $\lambda$ be given by $\rho_\lambda(S,t)$. The search proceeds from the top layer $\lambda = L$ to the earliest layer $\lambda = 1$. Then, if the search is proceeding successfully, for each network layer, $\rho_\lambda(S, t+1) \geq \rho_\lambda(S, t)$. There is no optimal value here only a constraint that must be satisfied over the time interval of the localization action. Constraints are often used in optimization problems to delimit the space of feasible solutions. In ST, this also plays the role of detecting when the constraint is violated and thus the search must be reset since the original *S* was incorrect. This is but one example of the likely many methods needed to detect errors in visual processing that must be corrected.

Thus, the strategy that will permit the system to maximize *S/(I+N)* is one where predictions are made as to what the signal of interest might be (either prior to a stimulus or after the first feedforward processing of the stimulus) and actions are taken to manipulate *I* so as to achieve the largest increase in the strength of the network responses corresponding to the prediction *S*. These actions include: priming the network for S by suppressing portions of the network that are irrelevant (when searching for one's daughter in the above example, the network units sensitive to image features related to walls, windows, doors, carpets, are likely irrelevant, among others) or imposing a surround suppression within the network for candidate neurons in order to reduce any local context effects.

These three examples illustrate the complexity of the control situation: it seems unlikely that any single objective function would suffice. The first, eye fixation change, is an easily recognizable objective function. The second, signal-to-interference-plus-noise ratio, is a very high level objective that refers to the overall system rather than one specific mechanism. The third example is not really an objective function; it is a constraint which if violated triggers a remedial action (namely, re-start the localization process on another focus of attention). There may be many such constraints in the overall system, each applicable for some particular action or situation.

## 6.0 Cognitive Programs and Attentional Mechanisms

As we have pointed out, the most basic process that we need to account for is comparison. To perform a comparison, one requires (at least): 2 things to compare, a comparison metric, a time at which to do the comparison, a way to actually perform the comparison, a way to represent the result, communication of the result to whatever other process needs it, and a way to reset the process for the next instance of comparison.

---

[8] A branch-and-bound algorithm is a common optimization method. It requires a representation that provides an enumeration of candidate solutions organized as a tree with the full set at the root. The algorithm explores branches of this tree, which represent subsets of the solution set. Conditions on an acceptable (or optimal) solution are checked for each branch, causing it to be discarded on failure or accepted otherwise (see Lawler & Wood 1966).
[9] This attentive surround suppression, spatial and featural, was predicted in Tsotsos (1990), and has been well documented since (e.g., Carrasco 2011).

We rely on Cognitive Programs to encode all of this including the appropriate timing information. The previous section discussed some aspects of control signals the CPs need to encode; this section will focus on examples of CPs and of Type II control signals.

An illustrative example of a CP appears in Tsotsos & Kruijne (2014), Macmillan and Creelman's (2005) Discrimination task. A graphical depiction of the cognitive program for a visual discrimination task is presented and the traversal of the graph from start to end provides the algorithm required to execute a discrimination task. The kinds of operations involve several instances of moving information from one place to another, executing a process, making a selection, or setting parameters. Specifically, this algorithm has the following steps:

  (i) the visual task executive receives the task specifications;
  (ii) using those specifications, the relevant CPs are retrieved from memory;
  (iii) the most appropriate CP is chosen and tuned into an executable script;
  (iv) the script is then executed, first activating in parallel the communication of the task information to the visual attention executive and initiating the attentive cycle;
  (v) the visual hierarchy is primed using task information (where possible) and in parallel attention is disengaged from the previous focus;
  (vi) the visual attention executive sets the parameters for executing the competition for selecting the focus of attention;
  (vii) disengage attention involves inhibiting the previously attended pathways and any previously applied surround suppression is also lifted (note that steps v–vii are executed in parallel before the visual stimulus appears, to the extent possible);
  (viii) the stimulus flows through the tuned visual hierarchy in a feed-forward manner;
  (ix) the central focus of attention is selected at the top of the visual hierarchy; the central focus of attention is communicated to the visual task executive that then matches it to the task requirements;
  (x) if the selected focus and the task requirements match, the task is complete.

The more complex examples of Visual Search are also presented there, a *1*-look version as well as a *n*-look version. These examples were useful but not proven that they would work if implemented, thus possibly incomplete as written. The next example is of a high performance test of the concepts.

STAR-RT is the first working prototype of Selective Tuning Attention Reference (STAR) model and Cognitive Programs (CPs). To test STAR in a realistic context, we implemented the necessary components of STAR and designed CPs for playing two closed-source video games, Canabalt and Robot Unicorn Attack (see Kotseruba & Tsotsos, 2017, for pointers to these games and other details). Since both games run in a browser window, our algorithm has the same amount of information and the same amount of time to react to the events on the screen as a human player would. STAR-RT plays both games in real time using only visual input and achieves scores comparable to human expert players. Specifically, all efforts were made to encode a single CP that could parameterized for both games. It thus provides an existence proof for the utility of the particular CP structure and primitives used and the potential for continued experimentation and verification of their utility in broader scenarios.

The Cognitive Program that controls the Canabalt game is graphically depicted as a high level algorithm in Figure 4 (from Kotseruba & Tsotsos 2017). It is comprised of a sequence of behaviors (move gaze to the left, release the button, etc.), decisions with yes-no results, perceptual acts (is drill on the roof? is runner detected? etc.), and working memory accesses (is button pressed? is timer expired?). The white boxes denote visual tasks such as detection, recognition, and spatial relations. The gray boxes denote non-visual elements, such a movement of gaze or hand actions. The connections between boxes can include the transfer of information, such as the position of the runner which is necessary to determine if he is close to the drill (which must be avoided). Working memory stores where the box was found or remembers if the button was pressed or not. Many attentional mechanisms are embedded here as well. Each change of gaze must be accompanied by a selection of new gaze point and a disengagement of attention from the previous fixation. Finally, the visual hierarchy itself must be capable of recognizing all of the relevant objects (runner, drill, rooftop, crate) and determining their locations with sufficient precision for actions (such as jumps). In the game, the runner is in constant motion and thus all decisions are time-critical in order to survive all the obstacles (also in motion). The CP in Figure 4 has been implemented in STAR-RT and tested on actual games. After playing 1000 games, the mean score of STAR-RT was over 3000m and its top score was 25,254 m, making it #18 in the all-time best human ranking (at the time). It is important to stress here that human-like performance scores were not our goal with this exercise. Rather, we wished to show that our concept of a Cognitive Program (as an extension and elaboration of Ullman's Visual Routines) is viable and leads to the desired functionality. The fact that the implementation performed so strongly was quite secondary (but still an impressive conclusion).

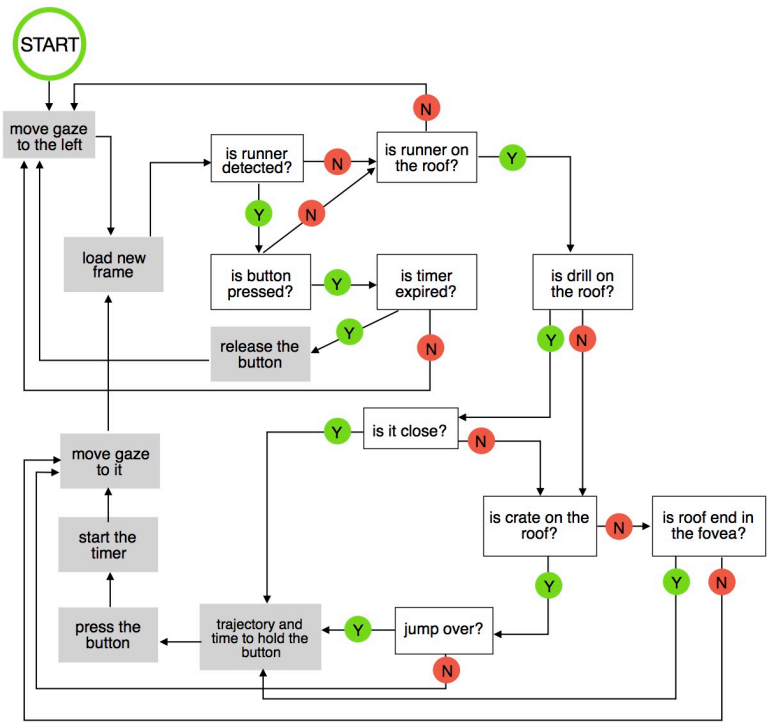

**Figure 4.** Canabalt Cognitive Program of STAR-RT. See text for details. (from Kotseruba & Tsotsos 2017).

A key idea underlying CPs as well as visual routines is that they be representative of a problem class and can be adaptable to similar situations. We tested this by adapting this CP to a second game of similar genre but visually very different (Robot Unicorn Attack). Few changes were required in order to make the system play a new game, specifically, the addition of visual processing routines to handle curved platforms, re-training the CNNs for different types of objects, and changing the control functions to add an extra key for dashing. Overall, the algorithm's style of playing both games is markedly different from human players. It makes more precise jumps than would be possible by using the physical keyboard, has better reaction time and is able to perform consecutive actions in quick succession. The important point is that, although not tested as extensively as the Canabalt CP, the Robot Unicorn Attack CP also performed very well demonstrating that the basic CP for this genre of gameplay was sound.

Figure 4 shows the CP as a flowchart of processes and decisions. Implied within this, is the fact that when the 'flow' proceeds from one box to the next, a 'move along' notice is needed. Something needs to inform the first box to terminate its processing, pass on any required information to the next stage, and instruct the second box initiate its processing. For example, the 'move gaze to it' box refers to moving gaze to the detected end of the roof. In a human agent this would mean extracting the location of the roof end, disengaging fixation (attention) from wherever it is currently, initiating an eye movement to the new location, ensuring the eye movement actually lands on the correct location (and correcting if not), and then engaging attention to that location. Other boxes refer to particular visual tasks. 'is runner detected' is an instance of Recognition. 'is runner on the roof' would require Recognition of the roof, localization of the previously detected runner as well as of the roof, and then a Spatial Relation computation for 'on'. Each of the steps of the CP would have similar sub-components.

A second implementation of CPs appears in Abid (2018). There he investigated how the basic elements of a CP might connect to a neural level realization. To do this, he used the neural motifs defined in Womelsdorf et al. (2014) for feedback inhibition, dendritic inhibition and disinhibition, feedforward excitation, and feedforward inhibition to accomplish the gating, integration and gain control computations the CPs for STAR require. He proposed that a basis set of elemental operations called Neural Primitives (NP) in conjunction with other control elements constitute a Cognitive Program. The Neural Primitives are biologically inspired computations that dictate the transformation functions from one representation to another and form the foundation of this thesis. With this was able to successfully simulate a number of classic attentional experiments from the literature, thus showing the plausibility of developing a theory that spans all three of Marr's levels of analysis.

Within each of the examples, there are implicit signals to step through the stages of the algorithms. Type II control signals seek to make those steps explicit. It is acknowledged that it might be that signal flow through sequences of neurons might accomplish this without any need for an explicit signal, but this is not yet clear. The reason we wish to make this explicit is to highlight how the various processes are inter-related and to functionally connect different aspects of solving a visual problem. Moreover, it points to possible experimental investigations that might determine whether or not such control signals exist, where they are needed or not, and their character.

Figure 5 depicts an example of how Type II control signals required for some of the visual tasks might appear in a CP. The only part of each signal that is shown is it's 'on' and 'off' point. The signals themselves appear in the middle portion of the figure, with the x-axis representing time. The way to parse this figure, in addition to reading the caption, follows. The time axis shown at the bottom applies to all levels of the figure. Part A is intended to show that control signals depend on the task of the moment. Each of the tasks considered are grouped by their combination of processing cycles on the visual hierarchy (following Tsotsos et al. 2008, Tsotsos 2011).

ST has different temporally ordered stages of visual processing (Tsotsos et al. 2008) beginning with top-down priming due to task instructions or knowledge and context, feedforward processing of visual signal, decision-making to confirm if a task is complete, recurrent processes to reduce network signal interference and permit identification of features or location, search of an image that would entail cycling through various parameterization of the previously listed tasks, and so on. The first group in Part A of the figure, Task and Knowledge Priming, represent the preparatory signals for the visual hierarchy that require a top-down pass of the visual hierarchy to set up in advance of the stimulus. The second group, Discrimination, Categorization, Recognition, need a single feedforward pass of the visual hierarchy to complete. The third group, Within-Category Identification, requires a feedforward pass followed by a partial top-down localization pass. The fourth group Localization and Gaze Shift require the feedforward pass and a full top-down pass. Finally, the last group, Visual Search, Compare, Measure, require multiple passes of the hierarchy in both directions. Each group is delineated by a colored line (visual search, for example, is the red line, while for the discrimination task it is the magenta line). The dashed portion of the task lines represent the pre-stimulus-onset period (where cueing might take place) and the solid part is the period when the stimulus is seen. The various components of Part B in the figure - priming, disengage, attention, etc. - are the specific signals that a control system must generate, with 'on' and 'off' shown at the appropriate times. These are strictly representative only.

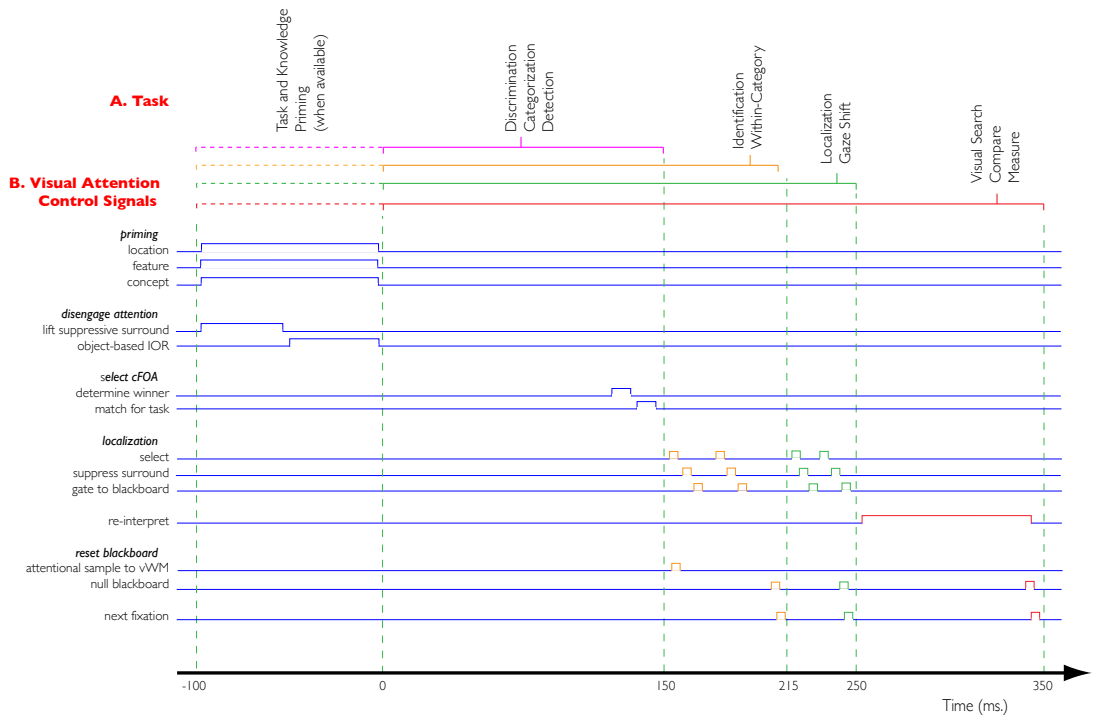

**Figure 5**. The control signals required for each of a number of different kinds of visual tasks. **A.** The set of tasks considered appear at the top, with the temporal extent of each shown in a different color. **B.** The set of control signals shown by the blues line to the right of the labels. These blue lines are intended to represent the timing of when the controller must generate an instruction appropriate to the indicated label. For example, in order to select the cFOA (central focus of attention), two actions are required at the end of the first feedforward pass through the visual hierarchy, namely, a selection of the strongest response followed by a matching function to compare what that strongest response represents to the goals of the task. Note how some signals (for example, 'null blackboard') have multiple and different colored pulses, one pulse for each layer of the visual hierarchy. (adapted from Tsotsos 2019).

The examples provided in this section demonstrate the plausibility of the overall control concepts and Cognitive Programs. However, they only suggest that the ideas represent how the brain actually performs such visual tasks and they do not detail how CPs may be formed or learned. These are issues for ongoing research.

Although many of the previous paragraphs concern particular methods and enumerations of algorithms and such, it would be incorrect to think that we advocate anything like the old *bag of 1000 tricks* approach to vision (Ramachandran, 1990; Purves and Lotto, 2003). Notably, Tsotsos (1993; 1995) argue against this and nothing in the present manuscript goes counter to those arguments. The bag of a 1000 tricks idea goes back to perhaps Selfridge's 1959 Pandemonium idea. There, Selfridge relied on 'demons', independent programs each specific to one particular input pattern. Some demons were specific for image features others played a higher level role (grouping, or decision-making). The key is that all demons operated in parallel at all times, each demon scanning all input to find its trigger pattern. Once triggered, it would compute a conclusion based on the input and add that to the input stream. In this way different patterns could be grouped into more abstract items. This is entirely bottom-up and is unable to consider any top-down information or local context or explicit controller. Our approach is nothing like this from the title onwards; our focus is on the role of attentional processes and their orchestration into a stable and predictable control strategy.

## 7.0 Discussion and Conclusions

We have shown how a computational approach to the problem of attentional control in vision might proceed. The contributions include: i) a Marr-inspired computational approach, from the general and abstract to the detailed and specific, considered as complementary and necessary for a complete understanding of the problem; ii) a reminder of the basic intractable nature of the vision problem when considered in its most generic form; iii) a suggested divide-and-conquer approach to developing solutions to vision sub-problems; iv) taxonomies of vision problems and of attentional mechanisms; v) arguments for why both a task executive and an attentive executive are necessary; vi) a Cognitive Programs strategy for encoding the algorithms that might solve these vision sub-problems; v)  computational level goal specifications of aspects of the attentional control problem; vi) a high level view of the kinds of timing signals that a CP must encode for a variety of basic vision tasks.

To this point, the reader would be forgiven to think that all of this is only a proposal; it is thus important to stress that most of the functionality has already been demonstrated, with a partial list being:
- vision hierarchy (e.g., Biparva & Tsotsos 2017, 2020, Tsotsos et al. 2005, Mehrani et al. 2020, Rodríguez-Sánchez & Tsotsos 2012);
- attention (Tsotsos 2011, Rosenfeld et al. 2018,  Bruce & Tsotsos 2009);
- fixation control (Wloka et al. 2018, Tsotsos et al. 2016);
- cognitive programs (Abid 2018, Kotseruba & Tsotsos 2017, Tsotsos & Kruijne 2014); and
- how ST, task and memory might interact (Berga et al. 2019, Tsotsos & Kruijne 2014).

As can be easily seen, the overall STAR architecture contains a full mix of stimulus, knowledge (memory) and tasks driven mechanisms. Certainly, there is much to do, but all of this provides a sound foundation.

Let us return to the question of why in this age of deep learning our computational approach for attentional control would not take advantage of its power. Several points must be made. First, we note that a computational approach can be more than simply simulating mathematics or building software for some narrow domain. There is nothing wrong with either, and often these can be very useful and important. But there is more. We can use the full breadth of computational tools as a foundation to help develop a mechanistic theory (in the sense of Brown 2014), something that many feel is lacking in the field. This is what this paper and the perspective it describes attempts to do.

Second a learning approach assumes that a sufficient corpus of the relevant data for training can be obtained. It also assumes that the 'answer' to some question actually lies within the statistical characteristics of the data. One would need to know both the question and what an answer might look like in advance. Considering our 3D active observer Same-Different problem, it is far from clear how all the data internal to the observer can be obtained let alone millions of trials of it. Nor is it clear what the question is nor how to characterize  an acceptable answer. The very impressive systems we see coming our of the deep learning community simply do not apply. The recent success of Google's DeepMind with the StarCraft II game (Vinyals et al. 2019a, Vinyals et al. 2019b) is an example. The input-output behavior is known in advance. As the 40-person team writes "In order to train AlphaStar, we built a highly scalable distributed training setup using Google's v3 TPUs that supports a population of agents learning from many thousands of parallel instances of StarCraft II. The AlphaStar league was run for 14 days, using 16 TPUs for each agent. During training, each agent experienced up to 200 years of real-time StarCraft play. The final AlphaStar agent consists of the components of the Nash distribution of the league - in other words, the most effective mixture of strategies that have been discovered - that run on a single desktop GPU." Even though the final version is relatively compact, getting there is not practical nor could it provide any kind of insight into how a human learns to play the game. This feat relies on a game engine that can generate data endlessly; our domain of interest has no such engine either for the actions observable on the exterior of a human agent nor for the internal and thus unobservable workings of cognition.

Third, the dominant approach in deep learning systems regarding attention is to use attention as equivalent to salience in a manner that seems to implement spatial attention in an early selection strategy (e.g., Vaswani et al. 2017). In fact, salience-driven early selection within feedforward vision has been shown to not represent human perception at all (Tsotsos et al. 2019). Although there have been successful integrations of the mechanisms of ST into CNN and deep learning systems (e.g., Biparva & Tsotsos 2017, 2020, Zhang et 2018, Gu et al. 2018, Roy & Todorovic 2017, Cornia et al. 2018), we did not take that route here because we are not interested in simply building a system to mimic human behavior on a visual task, such as Same-Different. Rather, we wish to provide an explanation for how humans perform that task and not a black box system that mirrors input-output behavior.

Fourth, it seems that both past experimental work as well as deep learning approaches in any domain neglect one important aspect: how is progress towards a problem's solution monitored during its execution, how are deviations detected, and how are are such unplanned failures corrected. Recall the dance workshop example from above where this is central. It seems clear that in order to monitor the progress of an algorithm's execution, the ability to inspect some or all intermediate stages and representations is necessary. The intermediate layers of representation found in deep learning systems do not easily lend themselves to this. Such a blanket statement is easily challenged; there is some evidence on inspection of neurons in early layers and many are active in trying to better understand the representations. However, it is not at all straightforward as to how to use the distributed representations that pervade network architectures in the localized fashion we require. This is not to say that this can never be solved; perhaps it will, but for now, such representations do not seem to have the right character for the issues of plan monitoring and repair.

We thus leave learning issues for future consideration once we have a much better understanding of what needs to be learned; a complete description of the problem at Marr's computational level will help provide this.

In summary, many experimental research efforts have provided a variety of insights into the kinds of brain structures that play a role in visual attention and its control. As the review in Section 2.0 shows, many important ideas as well as computational explanations have been proposed. In a sense, these are like the tendrils of a climbing plant reaching ever upwards but with no obvious target to spiral around as support. Our approach is an attempt to provide that target and support, by specifying the nature of the control problem that the brain solves. The downward computational branches we have described will seek to connect to the upwards tendrils with the hope that eventually they form a full theory. In other words, we feel that our approach is both complementary and necessary to the bottom-up experimental methods.

By way of examples of this interplay, we note the several experimental directions spawned by the original Selective Tuning papers that have now led to new knowledge (summarized in Tsotsos 2011, Carrasco 2011, Hopf 2018, among many). For our control concepts, Tsotsos & Womesldorf (2016) laid out a few supporting studies as well as new predictions. Primarily, they outline, as does the current manuscript in the context of the Selective Tuning model, how attentional control signals and their timing depend on the specific task being performed. They claim that tasks define sequences of computations with unique content and duration and predict that these computational sequences give rise to periodic, oscillatory activity that is measured across visual and fronto-parietal networks implementing attentional control of vision. These assertions can be tested. They argue that the computational constraints underlying visual tasks suggest a temporally precise unfolding of neural activation that is likely evident in brief periods of oscillatory activity measured across visual and attention networks of the brain. To be sure, we have a very long way to go, but the computational first principles approach we show here has significant promise.

**Acknowledgements:** This research was supported by several sources for which the authors are grateful: Air Force Office of Scientific Research (FA9550-18-1-0054), the Canada Research Chairs Program (950-231659), and the Natural Sciences and Engineering Research Council of Canada (RGPIN-2016-05352). No funding source had any role in the plan or performance of this research nor in the preparation of this paper.